\documentclass[10pt, conference, letterpaper]{IEEEtran}
\usepackage{cite}
\usepackage{amsmath,amssymb,amsfonts}
\usepackage{algorithmic}
\usepackage{graphicx}
\usepackage{textcomp}
\usepackage{xcolor}
\usepackage[linesnumbered,ruled]{algorithm2e}

\usepackage{flushend}

\usepackage[bookmarks=false,implicit=false]{hyperref}

\usepackage{array}
\def\BibTeX{{\rm B\kern-.05em{\sc i\kern-.025em b}\kern-.08em
    T\kern-.1667em\lower.7ex\hbox{E}\kern-.125emX}}
\begin{document}

\title{Privacy-Preserving Learning of Human Activity Predictors in Smart Environments
}

\author{\IEEEauthorblockN{Sharare Zehtabian, Siavash Khodadadeh, Ladislau B{\"o}l{\"o}ni and Damla Turgut}
\IEEEauthorblockA{\textit{Dept. of Computer Science, University of Central Florida, Orlando, FL, USA} \\
% \textit{University of Central Florida, Orlando, FL, USA}\\
 \\
sharare.zehtabian@knights.ucf.edu, siavash.khodadadeh@knights.ucf.edu, Ladislau.Boloni@ucf.edu, Damla.Turgut@ucf.edu}

}

% packing more
\renewcommand\floatpagefraction{.9}
\renewcommand\topfraction{.9}
\renewcommand\bottomfraction{.9}
\renewcommand\textfraction{.1}
\setcounter{totalnumber}{50}
\setcounter{topnumber}{50}
\setcounter{bottomnumber}{50}

\newcommand{\BL}[1]{{\color{brown} {\bf Lotzi:} #1}}
\newcommand{\SZ}[1]{{\color{red} {\bf Sharare:} #1}}
\newcommand{\SK}[1]{{\color{blue} {\bf Siavash:} #1}}
\newcommand{\DT}[1]{{\color{green} {\bf Damla:} #1}}
% adds page number remove before submitting
% 9 pages + references
%\pagestyle{plain}

\maketitle

\begin{abstract}

The daily activities performed by a disabled or elderly person can be monitored by a smart environment, and the acquired data can be used to learn a predictive model of user behavior. To speed up the learning, several researchers designed collaborative learning systems that use data from multiple users. However, disclosing the daily activities of an elderly or disabled user raises privacy concerns. 

In this paper, we use state-of-the-art deep neural network-based techniques to learn predictive human activity models in the local, centralized, and federated learning settings. A novel aspect of our work is that we carefully track the temporal evolution of the data available to the learner and the data shared by the user. In contrast to previous work where users shared all their data with the centralized learner, we consider users that aim to preserve their privacy. Thus, they choose between approaches in order to achieve their goals of predictive accuracy while minimizing the shared data. To help users make decisions before disclosing any data, we use machine learning to predict the degree to which a user would benefit from collaborative learning. We validate our approaches on real-world data.

\end{abstract}

\section{Introduction}
\label{introduction}

In the last two decades, a large number of research and commercial projects developed assistive technologies that aim to improve the quality of life for elderly or  disabled individuals. These systems are denoted by terms such as smart environments or smart homes for the elderly, ambient assistive living systems\cite{rashidi2012survey,acampora2013survey} or internet-of-things (IoT) assistive technologies. To choose the appropriate assistive actions, smart environments monitor and log the user's activities of daily living. One way to use this data is to learn a {\em predictive model} of the human activities~\cite{wu2017survey} $M(a_1...a_t, E) \rightarrow p(a_{t+1})$ which predicts the probability of the
next action $a_{t+1}$ given the actions performed by the user up to this point in the day and the parameters of the environment $E$ that includes the time of the day, day of the week, the weather and other information. A predictive model allows the system to anticipate the needs of the user and, furthermore, allows the system to detect short and long term changes in the user behavior. For instance, the environment might prompt the user to take his or her medication~\cite{seelye2012application} if he/she forgot to do so. Long term changes in the patterns of daily living might indicate a change in the underlying health condition, which need to be brought to the attention of caregivers~\cite{debes2016monitoring, ghayvat2019smart}. 

How does the smart environment learn the predictive model? Machine learning, in particular deep learning models, made significant progress in the last decade. However, many deep learning algorithms work best under {\em big data} regimes, where the number of data samples are counted from the tens of thousands (e.g. the MNIST dataset) to 500 billion (the Common Crawl dataset used to train the GPT-3 model). The activity logs of assistive environments, in contrast, are an example of {\em small data:} the number of individual activities performed by a user each day is counted in dozens, and we expect the model to yield actionable predictions in a matter of weeks after the system deployment. 

A possible solution to this dilemma is the use of {\em collaborative learning} which, by building a common model $M_\textit{shared}$ from the data of a pool of users, operate closer to the big data regimes favored by deep learning algorithms. The simplest choice of collaborative learning is {\em centralized learning}: the environments  transfer their logs to a cloud-based central authority, which combines these logs into a common training set. A different variant of collaborative learning, {\em federated learning}~\cite{konevcny2016federated}, also relies on a cloud-based central authority but requires the environments to perform learning locally and transfer only parameters of the learned model to the central system. Having access, directly or indirectly, to more data, collaborative learning promises faster convergence. 

An assumption of the collaborative learning approach is that the logs used for training are independently and identically distributed (iid). The daily routines of different users have clearly much in common due to biological and cultural factors as well as medical recommendations. On the other hand, every user has his/her own preferences, and the nature of the environment, the home and surroundings might also affect the schedule. For instance, a morning walk that is feasible in California in January might not be feasible in Minnesota. The fact that the iid assumption is only approximately satisfied bounds the performance achievable with collaborative learning. Local learning, which uses only the data collected from the given user is not subject to this limitation, as we can assume that these data is iid (at least over timespans that does not include significant lifestyle changes). 

A very important aspect of learning in smart environments is the consideration of {\em privacy}. The elderly and disabled are a vulnerable population, frequently targeted by hackers and scammers. Furthermore, the benefits of a smart environment are contingent on the trust of the user, which is strongly correlated with privacy. One of the fundamental principles of privacy is that of {\em data minimization}. In the context of  machine learning this principle means that the minimum amount of training data must be collected from users in order to acquire the specific benefits of the application. This  principle was stated, among others, in the consumer privacy report of the US White House in 2012~\cite{whitehouse2012consumer}, by the UK Information Commisioner's office~\cite{DataMinimization-UK} and it is also embedded in the European Union's General Data Protection Regulation (GDPR)\cite{DataMinimization-EU}. 

The principle of data minimization, applied to a smart environment means that the environment should not disclose information unless it provides a quantifiable benefit. When this is not feasible, and everything else being equal, the system should prefer techniques such as federated learning, which can be used in ways to achieve differential privacy~\cite{geyer2017differentially} to techniques such as centralized learning where privacy depends on assumptions about the central authority. However, the choice is not clear cut:  attacks against federated learning systems had been recently demonstrated~\cite{bagdasaryan2020backdoor}, and even in the absence of an attack information can still leak through, sometimes simply by the participation of a home in a given federated learning pool. Finally, we expect centralized learning to build a shared model better/faster than federated learning (as all the information available the the federated model is also available to the centralized one, but not the other way around). Whether the gap between these models is significant can only be evaluated through experiments with real world data. 

\bigskip

 We compare and track in time the evolution of the performance of the different approaches on a user-by-user basis. The main contributions of the paper are as follows:

\begin{itemize}
    \item In this paper, we train state-of-the-art deep learning based activity predictors using local, centralized and federated learning, using real world data from the CASAS dataset~\cite{cook2010learning}. 
    \item In contrast to previous approaches, we consider privacy conscious users/environments that are not sharing data if no benefit accrues from it. With this assumption, we carefully track the data available to different training approaches at specific points in time and measure the impact of the variation of training data on the prediction performance for different users. 
    \item We outline a machine learning technique through which a user can predict if it will benefit from the participation in a collaborative learning model {\em before} it shared any data.  
\end{itemize}

\section{Related Work}
\label{relatedwork}

% \BL{This is the paper that Diane Cook cites as introducing CASAS, so this is the one we need to cite} CASAS project~\cite{rashidi2009keeping} 

% Generally, recognizing patterns in residents' activities in a home and obtaining knowledge about the home's physical setting provides various advantages for the residents. For example, improving the comfort and safety of the residents and optimizing resource consumption. To this end, having a system to collect and analyse the physical and behavioral data is required. 

The contributions of this paper have two distinct intellectual lineages. As an application area, our topic is part of the field of human activity modeling and prediction. A significant subset of this work had been done in the context of smart environments, which provide both training data and an opportunity to use the models for the social good. From the perspective of artificial intelligence and machine learning, our work has its origin in collaborative learning techniques that aim to learn models while preserving the privacy of the sources of data. Federated learning is a recent and highly impactful technique in the field. 

\medskip

\noindent \textbf{Predicting future events in smart environments.} Most of the research in prediction in smart environments can be grouped into two categories: predicting the activities of daily living and predicting the location of events~\cite{wu2017survey}. 
% Dixit and Naik~\cite{dixit2014use} 

The development of machine learning based approaches for such predictors require training data which is considerably harder to obtain for a human-inhabited physical system such as a smart-home compared to domains where training data can be simply obtained by scraping the internet. The CASAS dataset~\cite{rashidi2009keeping} is one of the most complete, maintained and publicly available smarthome datasets; it had been the catalyst of many subsequent research efforts.

%Rashidi and Cook~\cite{rashidi2009keeping} introduced CASAS smart home system that collects sensor data and obtains useful knowledge from this data by using machine learning and data mining techniques. 

Minor, Doppa and Cook~\cite{minor2015data} used the CASAS dataset to learn activity predictors. In contrast to our predictor which predicts the probability of the next event (essentially, a probabilistic classification problem), this work predicts the individual time delays when the next event of a given class would take place (a regression problem). To this end, the authors trained separate regression models with model trees of each class. Another work that focused on a current activity recognition task based on the sensory data inputs in CASAS datasets is by Liciotti et al. ~\cite{liciotti2020sequential}.

%focused on current activity recognition task based on sensory data inputs in CASAS datasets. 

% they assume that all the activities will take place in a given day, and predict the next occurrence of the 

%\BL{Does that paper talk about activity prediction? \SZ{Yes, but their approach is mostly data mining. If I understood correctly, they find repetitive sequence of activities and the periodicity between those sequences (for example every 3 hours). Then they recognize startup trigger action for each activity. when that action happens the system can predict the activity. }}

Mshali et al.~\cite{mshali2018adaptive} developed an e-health monitoring framework to detect abnormal and risky daily activities and predict the health conditions of the residents using a Grey prediction model (GM)~\cite{kayacan2010grey}.
Choi et al.~\cite{choi2013human} proposed two deep learning algorithms based on deep belief networks~\cite{le2010deep} and restricted Boltzman machines~\cite{larochelle2008classification} to predict the behaviors of residents using MIT home dataset~\cite{tapia2004activity}.

% They take the sensory data from the previous 45 minutes and predict which sensors will be activated in the next 5 minutes.
% The states of sensors in the next few minutes can be very similar to the states of them in the current time. 
%While activity recognition requires processing small intervals (e.g. every 5 minutes), in this article, we particularly strive to predict longer-term and higher-level relations between activities by taking a sequence of activities to predict the next activity.

%This is used as foundation of our work. We are focusing on predicting the next activity based on a sequence of activities of the user.   

\medskip

\noindent \textbf{Federated learning.} Federated learning had initially been proposed as a technique to improve communication efficiency in distributed learning~\cite{konevcny2016federated}. However, it had been pointed out that the technique also allows the learning system to ensure differential privacy~\cite{geyer2017differentially}. One of the early, high profile applications was Google's Gboard~\cite{hard2018federated} which used federated averaging (FedAvg)~\cite{mcmahan2017communication} to improve next word prediction. In recent years, several research projects improved the performance and privacy characteristics of federated learning. Zhao et al.~\cite{zhao2018federated} suggested a data sharing approach to improve the performance of the FedAvg algorithm in case the training data is non-IID. Wang et al.~\cite{wang2019adaptive} aims to optimize learning of a gradient-descent based federated learning algorithm at the edge. In federated learning algorithms, local training happens at the edge and global aggregation is performed on a central place. They proposed a control algorithm that determines the best frequency of global aggregation with which computation and communication resources at the edge can be used efficiently in federated learning. Zhang et al.~\cite{Zhang2020Infocom} proposed building trustworthy federated learning systems using trusted execution environments (TEEs). Their main focus was to assure that the local training on clients side is being done correctly. 

% In order to achieve accuracy for smart home automation, sufficient and diverse data are required. Furthermore, to maintain the privacy of such data, a decentralized learning method is proposed. Federated learning has had successes in various applications. 

Yu et al.~\cite{yu2020learning} suggested a framework to automatically learn contextual access control policies for IoT devices in smart homes in order to detect if an access to an IoT device should be allowed or blocked. To learn an accurate model for this, sufficient and diverse data required which cannot be provided by a single home. On the other hand, collecting data from all smart homes will bring privacy concerns to the users. To address these issues, they leveraged federated multi-task learning (FMTL) approach. 
Nishio and Yonetani~\cite{nishio2019client} focused on client selection in federated training (FedCS). In their suggested protocol, the central mobile edge computing (MEC) operator sends a resource information request to the random clients. Based on their information such as computational capacity, wireless channel status and size of data, the MEC can decide which client will be able to participate in training and deliver the updates for global aggregation in time. For our client selection, however, we simulate the deployment time of smart homes and select the participants in training by that information.

% Zhao et al.~\cite{zhao2018federated} proposed a data-sharing strategy to improve FedAvg with non-IID data. They addressed the accuracy reduction by weight divergence challenge with creating a small subset of data which is globally shared between all the edge devices.

\section{Training Data for Collaborative Learning in Smart Environments}
\label{sec:Data}

The performance of machine learning models depends on the data used for training. Everything else being equal, more data is better, and highly expressive models, such as deep neural networks, require more training data to avoid overfitting. Many recent achievements of deep learning took place in a "big data" regime; Google, Facebook and Amazon rely on a steady stream of data from the users interacting with their services. For instance, a success story in federated learning is predicting the next word on a mobile device's keyboard~\cite{hard2018federated}, relying on a very large number of users receiving the collaborative learning client simultaneously. 

However, in the case of a smart environment participating in a collaborative learning scheme, this model cannot be taken for granted. A privacy conscious user (or the environment acting on her behalf) would not share any data unless there is a strong likelihood that it would benefit from the transaction. At the same time, the user will stop sharing when no further benefit is likely. As we have seen in the introduction, this data minimization behavior approach had been recommended by government directives in the US, UK and EU. As the problem of privacy is particularly acute for the vulnerable elderly and disabled population, the regulatory pressure is likely to increase. 

We need to emphasize that the data minimization principle does not preclude the use of collaborative learning and other cloud based techniques. It means however, that some of the simplifying assumptions are not applicable: we need a better understanding of the temporal aspect of data sharing: what training data is available, to whom and when. 

\bigskip

\begin{figure}
    \centering
    \includegraphics[width=0.5\linewidth]{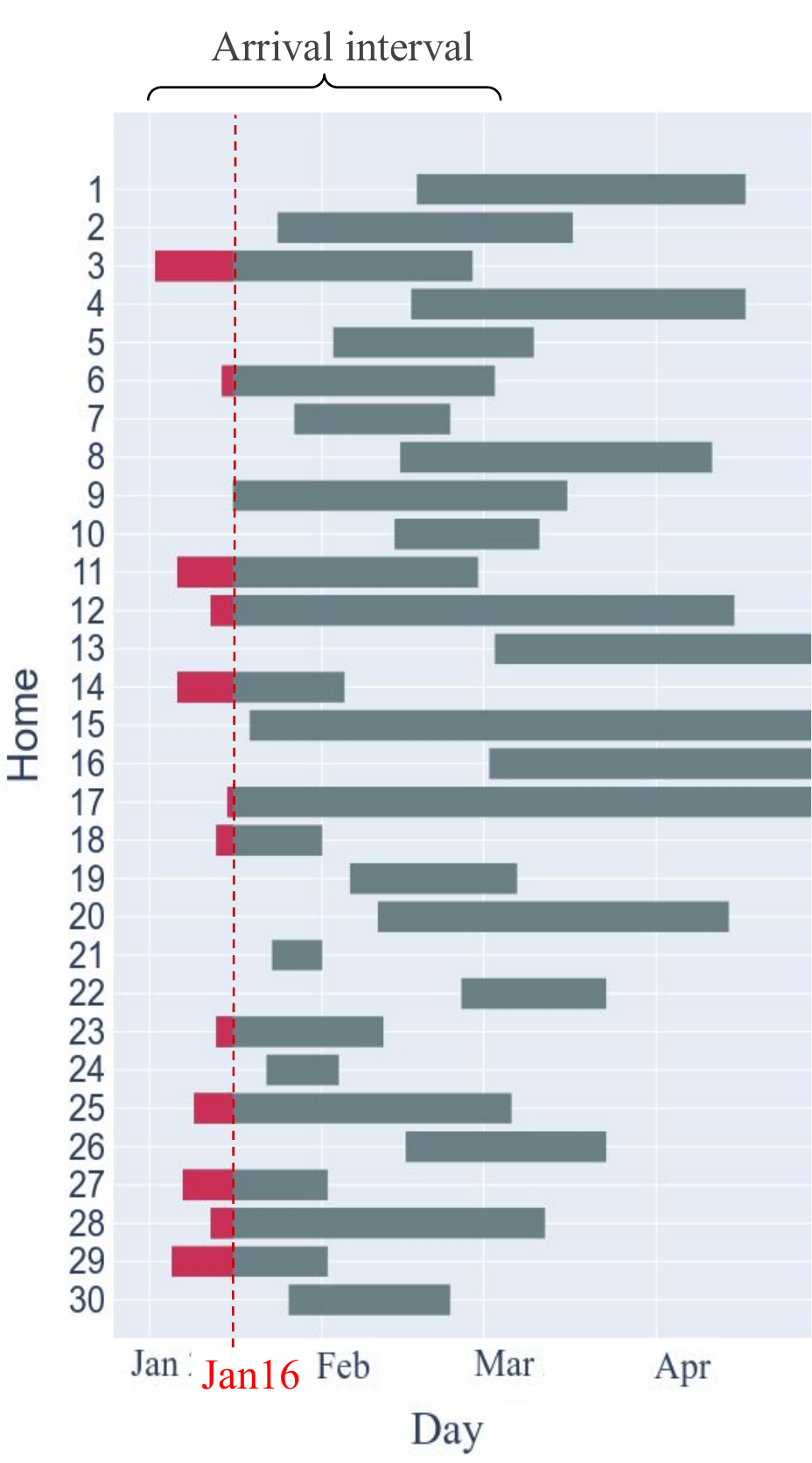}
    \caption{The data available for collaborative training for a group of users. The deployment time of the system is modeled through a Poisson arrival starting from January 1st. The users stop sharing data when further data sharing is not justified by the advantage of collaborative learning. The red part of the bars illustrates the data available for collaborative training on January 16th.}
    \label{fig:TrainingDataInTime}
    \vspace{-4mm}
\end{figure}

The first simplifying assumption we need to discard is the synchronized start of data collection. The deployment of a smart environment for the disabled or elderly is not instantaneous. It requires physical installation of hardware, software configuration, user training and possibly legal and medical approval. Thus, the smart environment will be deployed for some users earlier than for others. As the deployment times are random but independent from each other, they follow a Poisson arrival process. Furthermore, the number of smart environments contributing to a single collaborative domain is significantly smaller than the countrywide domains used by web services. Under these conditions, modeling the deployment time is necessary, because it affects the amount of training data available to the predictor. 

Let us discuss the problem of the data available for learning in a group of smart environments. We will consider a set of smart homes $H_1 \ldots H_M$ that started to operate at times $t^\textit{start}_i$ distributed according to a Poisson process with an arrival rate $\lambda$. We take the perspective of the target smart home $H_\textit{tg}$ that had started to operate on day $t^\textit{start}_\textit{tg}$. Figure~\ref{fig:TrainingDataInTime} illustrates an example with $M=30$, $\lambda=0.5$, starting time January 1st, and we are considering $H_\textit{tg} = H_\textit{14}$ with $t^\textit{start}_\textit{tg} =$ January 6.

{\em Local learning} involves the training of a model based on data collected from the same home. This approach has the highest level of privacy, as no personal information needs to leave the premises. The weakness of the local training model is the paucity of the data, especially early in the deployment. On day  $t^\textit{start}_\textit{tg}$, the system has no training data whatsoever, on day  $t^\textit{start}_\textit{tg}+1$ it will only have one day of training data and so on. Thus we expect the accuracy to start from a very low level, but increase in time as training data accumulates. The red part of Home 14 data in Figure~\ref{fig:TrainingDataInTime} shows the data available to this home on January 16 in the local training regime (ten days of recordings from January 6 to January 15). 

Let us now consider the case of {\em centralized learning}. As the smart environment was deployed at different homes at different times, on the day the target home had started, a number of other homes are already operating and providing data. If home $H_i$ started at time $t^\textit{start}_\textit{tg}$, the total amount of training data available at that point will be: 
\begin{equation}
    D_\textit{tg}^\textit{centralized} = \bigcup_{i; t_i^\textit{start} < t^\textit{start}_\textit{tg}} D_i[t_i^\textit{start}:t^\textit{start}_\textit{tg}]
\end{equation}

For our example, the data available for training is the data from all homes where the system was deployed before January 16 - these are all the parts of bars shown in red in Figure~\ref{fig:TrainingDataInTime}. 

Another simplifying assumption that is not applicable for privacy-aware users is that once a user joined a collaborative learning setup, it will provide data indefinitely into the future. To do this might be in the interest of the central authority, but it is not compatible with the privacy principle of data minimization. A rational user will stop providing data to the central system as soon as the local learning yields better results than the predictive models received from the center. As shown by the termination of the bars in Figure~\ref{fig:TrainingDataInTime}, this {\em cross-over point} might happen sooner or later in time and it triggers the end of sharing data $t_i^{shend}$. We note that this time point only shows the end of data {\em sharing}; the smart environment will continue to operate and the local learning will continue to receive data past this time. Thus the data available for centralized learning will be:
\begin{equation}
    D_\textit{tg}^\textit{centralized} = \bigcup_{i; t_i^\textit{start} < t^\textit{start}_\textit{tg}} D_i[t_i^\textit{start}:\min(t^\textit{shend}_i, t^\textit{start}_\textit{tg})]
\label{eq:training-data-centralized}
\end{equation}

{\em Federated learning} operates on the same amount of data, with the difference that the data is never put together to a shared database.

\section{Learning the Activity Prediction Model}
\label{method}

In this section we describe the architecture and training process of a human activity predictor for a smart environment that predicts the future activities of the residents based on the history of activities and current environment. We represent the input as a sequence of tuples $(h, d, a)$ containing one-hot encoded hour of the day $h$, day of the week $d$ and activity label $a$. Our predictor $f$ takes a sequence of $l$ tuples and outputs the probability of occurrence of next activity:
\begin{equation}
    f((h_{t}, d_{t}, a_{t}), ..., (h_{t + l - 1}, d_{t + l - 1}, a_{t + l - 1})) \xrightarrow[]{} p(a_{t+l})
\label{eq:predictor}
\end{equation}

Our goal is to find the ``best'' predictor. One way to formalize this is by assuming that the function $f$ is part of a parameterized family of sufficiently expressive functions $f(\cdot)=F(\cdot, \theta)$.
% {\color{blue} In other words, the function has a sufficient amount of parameters to predict based on trends in data.}
In our case, this family will be a particular type of deep neural network, and $\theta$ will map to the network weights - but many other choices exist. Thus finding the best function is mapped to finding the optimal $\theta = \theta^*$. 

Naturally, we cannot exactly predict every activity due to the inherent randomness of the human behavior. We will define the accuracy of predictor in the form of a loss function expressed as the cross-entropy between the predicted probabilities and the actually occurring activity. The optimal $\theta^*$ will be the value that minimizes this loss over the available training data. In the remainder of this section, our focus will be on finding the appropriate form for the function $F$ and the optimization process for finding $\theta^*$.

\subsection{A Long-Short Term Memory Based Activity Predictor}

\begin{figure}
    \centering
    \includegraphics[width=\linewidth]{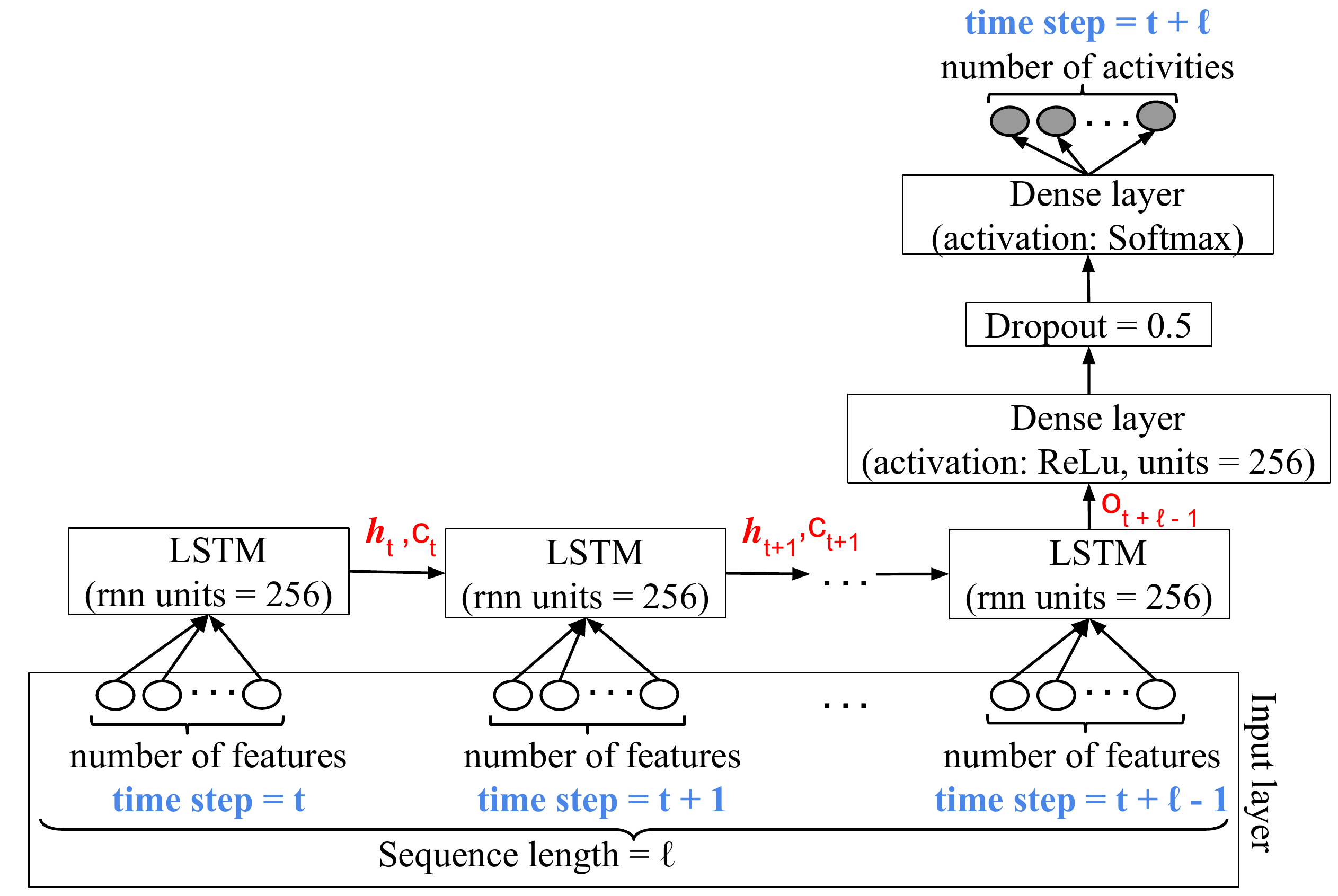}
    \caption{Architecture of LSTM based prediction model. Each circle in the input layer shows a single feature. A set of features is considered as the inputs in each time step. Gray circles in the output of the model correspond to the activities. 
    $h$ is the hidden state and $c$ is the cell state. 
    % After processing the whole sequence, the output $o$ is fed to a fully connected layer with 256 units.
    }
    \label{fig:lstm}
\end{figure}

In recent years, time series predictors based on a specific type of recurrent neural networks, Long-Short Term Memory (LSTM)~\cite{hochreiter1997long} had been successfully applied to problems ranging from natural language processing~\cite{wang2016learning, ghosh2016contextual} to robotics, computer vision and taxi demand prediction~\cite{chalvatzaki2019lstm, Kong2018ActionPF, xu2017real} and predictive caching~\cite{Zehtabian-2021-PMC}. Compared to other machine learning approaches where feature engineering is essential, deep neural networks, trained end-to-end using stochastic gradient descent, learn their own latent feature encoding. Within the field of deep neural networks, LSTMs have the advantage of having a learned memory state. This allows a prediction to be conditioned on events that happened many time steps in the past, while still handling one event at a time.

Fig.~\ref{fig:lstm} shows the architecture of a deep neural network designed to learn the prediction function Eq~\ref{eq:predictor}. The input layer of shape $l \times n$ encodes the $l$ tuples of history. The second layer is an LSTM of size 256 unrolled $l$ times. The hidden state $h$ and the cell state $c$ (memory) in the previous time step alongside the input in the current time step is given to the current LSTM cell. This procedure runs repeatedly until all data in the given sequence is processed. At that point, the output of the LSTM cell $o$ will become the input of the next layer, which is a dense layer with a ReLU activation function. This layer is followed by a dropout layer~\cite{srivastava2014dropout} with a dropout rate of 0.5 to improve the generalization of the model. Finally, we have another dense layer with a softmax activation function that outputs a probability for each activity. For training purposes, we are using a cross-entropy loss between the output of the softmax and the actual next activity. When deployed and used as a predictor, the smart environment can take the activity with the highest probability to be the predicted activity for the next time step. 

In the following, we describe three possible scenarios for training activity predictors for the smart environments: local training and two collaborative training scenarios - centralized and federated. As we predict activities in real-time since after training, we only need to do a feed forward pass to compute the predictions. Following the practice of deep learning literature, to make fair comparisons for all the training models we are using exactly the same predictor architecture from Fig.~\ref{fig:lstm}.

% For all the training models, we are using exactly the same predictor architecture from Fig.~\ref{fig:lstm}.

\begin{figure}
    \centering
    \includegraphics[width=0.95\linewidth]{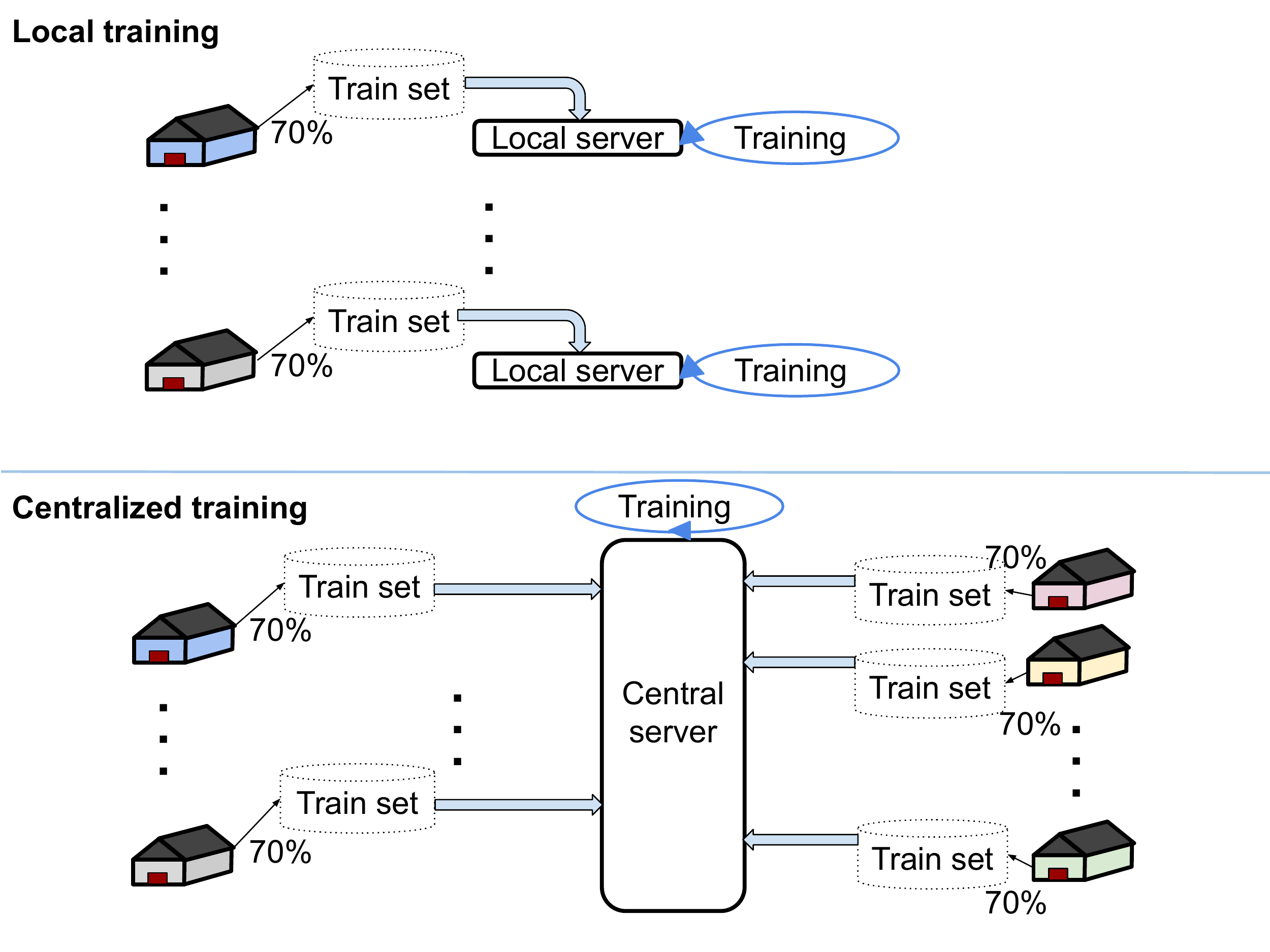}
    \includegraphics[width=0.95\linewidth]{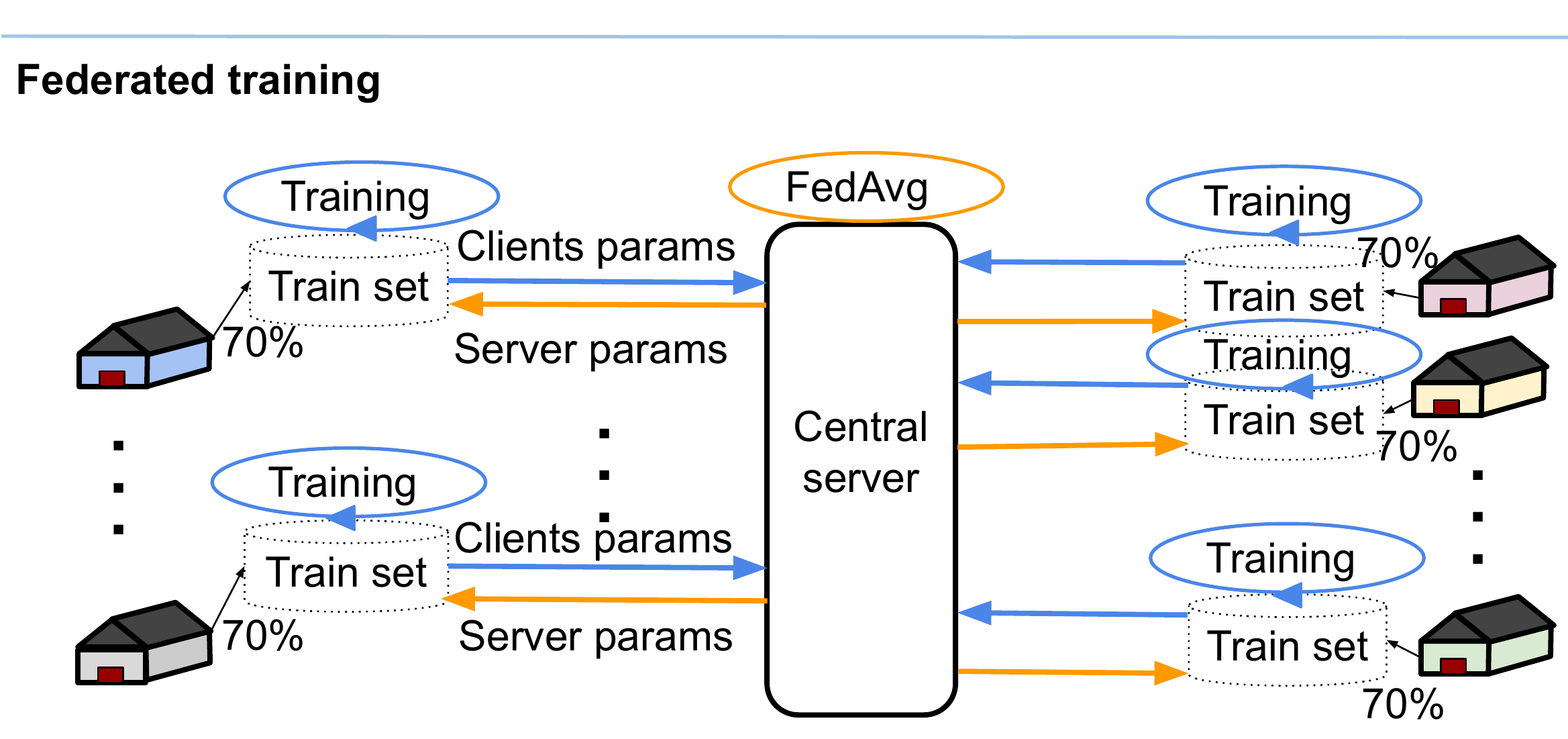}
    \caption{Activity prediction approaches 1. local (in-home) training (top), 2. centralized training (middle), and 3. federated training (bottom). We used 70\% of each home's data for training and 30\% for testing.
    }
    \label{fig:methods}
\end{figure}

\subsection{Local Training}

For the case of local training, the smart environment uses only the data collected from the user to train a predictor specific for the environment. Overall the number of trained predictors is the number of deployed environments. The advantage of this model is that the predictor is tailored to the home. The disadvantage is that the network is training with less data  for instance, for the first day there is only one day of training data. The training process will be repeated overnight using the full set of data available for the node. The local training process is outlined in Fig~\ref{fig:methods}-Top.

The accuracy of the prediction is measured on the home's own data. In general, we expect the paucity of the training data to result in an initially weak predictor which, however, will improve in time as more training data becomes available.

% One approach for predicting the residents activities is to train a model locally based on the data collected earlier in the same home. For example, we can have a predictive model in a local device that gets updated with a new data. Figure~\ref{fig:methods} top, shows the overall layout of local training approach. Each home in this architecture has its predictive training model. We split the whole dataset for each home into training data and test data with a 70\% and 30\% ratio, respectively. Then we feed the training data into the local prediction model in a daily basis and after each day of training, we evaluate the prediction on the test data. This means that the size of the training data in the $k$th day is equal to $k$ days of the data. Note that the size of the number of data points of $k$ days for one home is not necessarily the same as other homes'. If number of collected data for one home is less than defined sequence length, we skip that home in that day of training. 
% We use $M$  LSTM predictive models that $M$ is the number of smart homes.

\subsection{Centralized Training}

In the case of the centralized training model we assume a cloud-based central authority. The participating homes upload their daily logs to the central authority as training data. The central authority runs the learning algorithm daily, creating a single predictor which is transferred to the homes. The training data available to the centralized learning is described by Eq.\ref{eq:training-data-centralized}. This learning model is shown in Fig~\ref{fig:methods}-Middle. 

The accuracy of this predictor is then evaluated on the local home's test data. Thus, the same predictor will have different accuracy results in different homes. A positive aspect of the centralized predictions is that the learning happens with much more data especially compared to the local learning for smart environments recently deployed. A weakness of the model is that the centralized learner learns a shared model, and will not adapt to the preferences and idiosyncrasies of the individual users. Users that joined the centralized learning pool will provide more data and they had more opportunities to shape the predictors to their own routines. We thus expect that for each node the centralized learning model will start with a better accuracy for a newly joined node, but it will improve comparatively slower from there.

\subsection{Federated Training}

Federated training is a variant of collaborative learning which does not require the participating nodes to share their data. Each node implements a learner that has access to the locally generated data. As in the centralized training approach, there is a cloud based central authority that learns and distributes a shared model. However, in contrast to the centralized approach, the central authority does not receive training data from the environments, but parameters from the locally updated models. 

There are several techniques through which the federated learner can update the shared model. The approach we use is the federated averaging model introduced by McMahan et al.~\cite{mcmahan2017communication}, due to its robustness to imbalanced datasets like the ones found in smart environments with different deployment dates blue (see Fig~\ref{fig:stats}), where some homes provide significantly more data as they started earlier.
The updated model is then transmitted to the nodes. Fig~\ref{fig:methods}-Bottom shows the organization of the federated training approach.

There are many similarities between the federated and centralized models. The total amount of data to which the system as a distributed learner has access is the same, as shown in Eq.\ref{eq:training-data-centralized}. The difference, however, is that the centralized model has access to this data directly, while the federated model only through the mediation of the parameters of the local learners.

The same considerations apply to the expected accuracy of the shared model. Everything else being equal, we expect the centralized approach to show a better accuracy than the federated one, because it has a better access to the training data. A way to illustrate this is that we can always emulate federated learning in a centralized fashion, but not the other way around. Naturally, due to the randomness inherent in human behavior, it is possible that for a given day the federated model to be better for a particular home than the centralized one. 

The expected weaker performance of the federated learning model is compensated by the better privacy properties. As the federated learning only shares parameters on the local model, it is expected that less information is disclosed compared to the centralized learning approach. 

The choice between federated and centralized learning for a privacy-aware agent boils down to the performance gap between them. If the performance gap is major, the smart environment is better off using a centralized approach (and, possibly, cutting off the data sharing when the local model catched up). If the centralized-federated performance gap is minor, the system is better off using federated learning.

\subsection{Predicting If Smart Environments Benefit from Federated Training}

Let us now summarize the expected accuracy profiles and the three learning approaches we consider:

\begin{itemize}
    \item Local training: will start with a low accuracy due to lack of training data, the performance will increase in time, and in principle is limited only by natural variability of activities and by model capacity. Privacy is guaranteed as no data leaves the premises. 
    \item Centralized training: will start with a higher accuracy due to existing training data from nodes that were deployed earlier. The accuracy will increase relatively slowly and, in addition, will be limited by the non-iid distribution of the training data between different environments. Significant privacy concerns due to data sharing. 
    \item Federated training: accuracy profile expected to be similar to centralized training. Privacy concerns lower, but information leakage still possible.  
\end{itemize}

Note that we are expecting that eventually the local training will overtake the collaborative learning approaches. At this point, a rationally behaving privacy aware smart environment will stop participating in the collaborative learning model, stop sharing data and continue improving its activity predictor using local learning. 

One additional insight we must consider is that simply participating in the collaborative learning and sharing a single day's activities might be the largest privacy loss, as it might disclose the user's age, medical needs and disability condition. Disclosing further day's data of the same daily routine will disclose relatively few additional information. Thus, the smart environment must consider carefully whether it should participate in the collaborative learning even for a short time. 

We are going to define a number of measurable quantities that would allow the environment to make these decisions. One such quantity is the {\em crossover point}: the day in the future from which the model acquired through local training will consistently overtake the one received from the collaborative learning (centralized or federated). Intuitively, the closer the crossover point is, the less justified is for the user to join the collaborative learning pool. 

The second quantity of interest is the area between the local and collaborative learning models accuracy in time up to the crossover point. Using a term borrowed from the field of reinforcement learning, we will call this quantity {\em regret} - this is the overall accuracy performance lost if the user does not join the collaborative learning pool. The smaller the regret, the less justified is to join the collaborative learning pool. 

Naturally, both the crossover point and the regret can be measured only after the fact. In this paper, we propose the hypothesis that while these values are difficult to predict, we can train a classifier for a good surrogate measure that can be used as a decision aid. We will create a classifier that, based on the histogram of the first $k$ days of the node and the average over all nodes will predict whether the crossover point will happen before specific day $d$ or not. 

As a note: training such a classifier requires the collaboration of the central authority, and might result in the node not joining the collaborative learning pool. Thus, it would not be in interest of the centralized authority to provide this classifier if the authority has a business model that relies on data sharing. However, it {\em would} be in the interest of the authority to help make this decision if the privacy interests of the central authority and the nodes are aligned.

\section{Experimental Study}
\label{experiments}

\begin{figure*}
    \centering
    \includegraphics[width=0.97\textwidth]{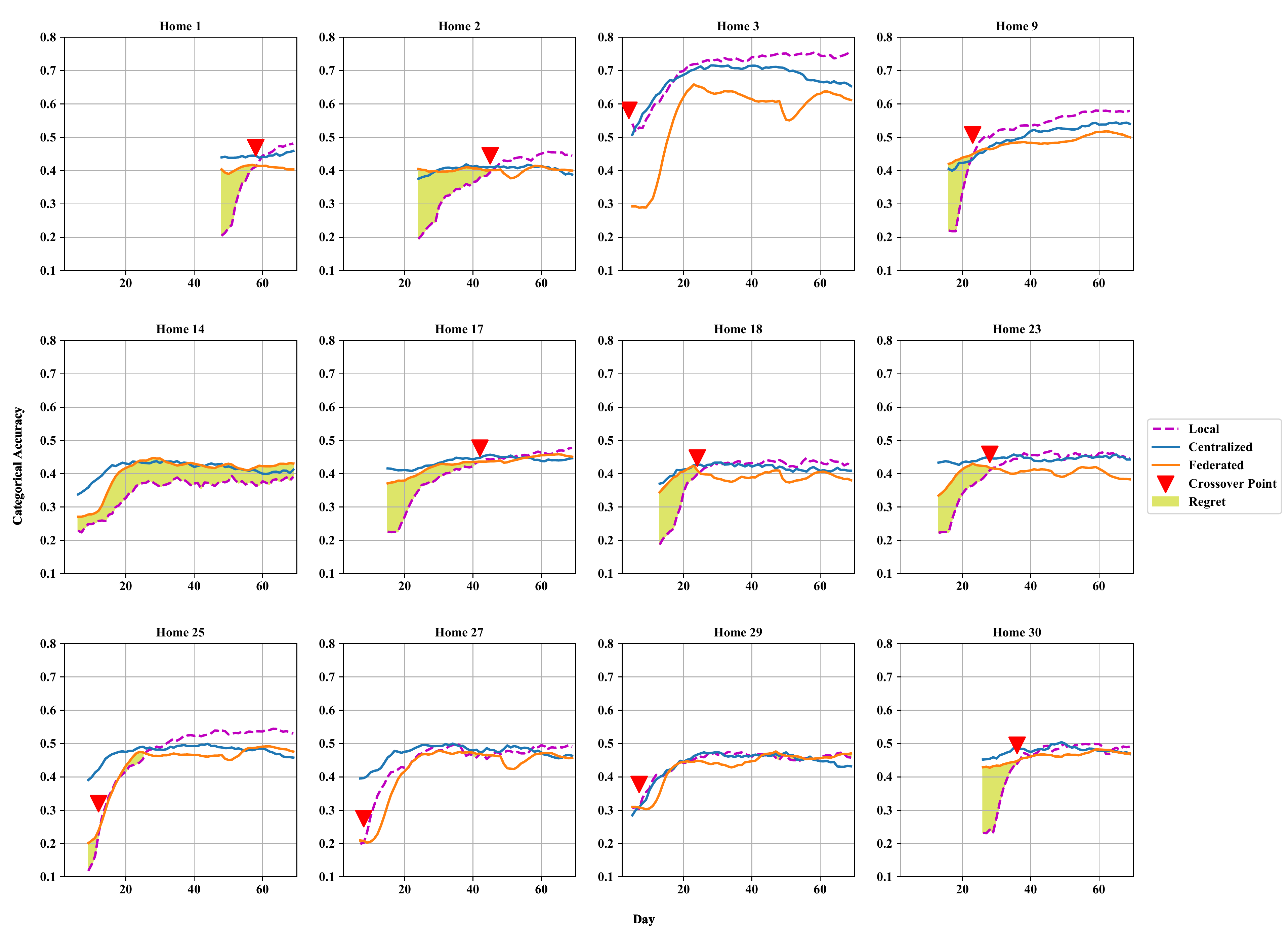}
    \vspace{-2mm}
    \caption{Accuracy on test data for a selection of 12 out of the 30 homes in our dataset with local training (magenta) vs centralized training (blue) and federated training (orange). The cross-point shows the first time that the local accuracy reaches the federated accuracy. Regret is the area between local accuracy and federated accuracy when local accuracy is lower. than federated accuracy.}
    \vspace{-3mm}
    \label{fig:evaluation}
\end{figure*}

In the previous section we made certain conjectures about the accuracy profiles of the activity predictors. {\em Qualitatively}, these predictions are supported by objective facts: we know that local training has less training data than collaborative ones, and we know that centralized training can emulate federated learning but not the vice-versa. However, any practical application would need to rely on the {\em quantitative} results. For instance, if the crossover point would take years to reach, collaborative learning would be the only reasonable alternative for a smart environment. If the difference between the centralized and federated learning results is large, the system will need to choose centralized learning even if privacy vulnerabilities exist. 

These quantitative factors strongly depend on the actual data. We could be right about the overall patterns, but wrong about the scales at which these patterns happen. Performing experiments using real world data is the only way in which we can understand the decisions faced by smart environments. 

\subsection{Datasets and Pre-processing}

For our experiments we used the datasets collected by the CASAS project~\cite{cook2010learning}~\footnote{
% \href{https://archive.ics.uci.edu/ml/datasets/Human+Activity+Recognition+from+Continuous+Ambient+Sensor+Data}
{https://archive.ics.uci.edu/ml/datasets/\\Human+Activity+Recognition+from+Continuous+Ambient+Sensor+Data}
}. This collection contains 30 datasets collected in homes with volunteer residents performing their daily routines. There is a significant diversity in the datasets and the routines: some of the residents were younger adults, some were healthy older adults, some were older adults with dementia, and some were having pets. 

In order to make the datasets suitable for our experiments, we performed several pre-processing steps. 

\bigskip

\noindent{\bf Mapping the activity labels into a common ontology.} The original activity labels are closely related, but not fully identical across the various datasets. Labels at various levels of granularity exist, such as {\em work}, {\em work at table}, {\em work on computer} and {\em work at desk}. The use of different labels would make any form of collaborative learning impossible, and local learning difficult to compare between datasets. We solved this problem by mapping the activity labels to a higher level, courser granularity categories, creating 10 new category labels without overlap between them. The mapping of the original activity labels to the new common ontology is shown in Table~\ref{tab:act_mapping}. 

\begin{table}
% \caption{Mapping the dataset activity labels to a higher level daily activity categories}
    \caption{Mapping the dataset activities to a higher level category}
    \centering
    \begin{tabular}{|>{\centering\arraybackslash}p{5.4cm}|>{\centering\arraybackslash}p{2.5cm}|}
    \hline
       \bf Original Activities  &  \bf Category\\
       \hline
       evening meds, morning meds, take medicine, exercise,  toilet, groom, dress, r2.dress, bathe, personal hygiene, r2.personal hygiene  & PERSONAL HEALTH AND HYGIENE\\
      \hline
       eat, eat breakfast, r2.eat breakfast, eat lunch, eat dinner & EAT \\
      \hline
       drink & DRINK\\
      \hline
        cook, r1.cook breakfast, cook lunch, cook dinner, cook, cook breakfast, wash dishes,  wash breakfast dishes,  wash lunch dishes,  wash dinner dishes, laundry & CHORES\\
      \hline
       nap, sleep, r1.sleep, sleep out of bed, go to sleep / wake up (interval between them) & REST \\
      \hline
       relax, watch TV, read & RELAX\\
      \hline
       phone, entertain guests & SOCIAL\\
      \hline
       work, work at table, work on computer, work at desk & WORK\\
      \hline
       leave home & LEAVE HOME\\
      \hline
       enter home & ENTER HOME\\
       \hline
       other activity, step out, bed toilet transition & NOT TRACKED  \\
       \hline
    \end{tabular}
    \label{tab:act_mapping}
    \vspace{-3mm}
\end{table}

%(Removed from the data)

\medskip

\noindent{\bf Converting to an event-based time series.} The CASAS dataset uses a variable sampling rate from a resolution ranging from several seconds to sometimes hours. When the sampling rate is fast, it usually results in many repeated entries with the same activity label. 

We have several choices to convert these entries into a format that is more suitable to machine learning. We could, for instance, map the entries into a shared, uniform time grid across all the datasets. However, this approach would create very large datasets, with redundant information. 

Instead, we chose to use an {\em event-based} representation of the activity time series, by representing every contiguous activity with a single entry. One side effect of this is that the dataset for each day will be significantly shorter, but entries for the individual days might have a varying length. This, however, is naturally handled by the LSTM-based activity recognition engine. Note that this approach does not encode the length of an activity through the number of repeated entities. However, the temporal information is still present by the encoding of the time of the day as one of the features. 

\medskip 

\noindent{\bf Feature representation} We are using a representation where every entry in the time series has three data fields: the hour of the day (as an integer 0-23), the day of the week (as an integer 0-6), and the activity label which is also encoded as an integer in the range 0-9. Each value was individually encoded with a one-hot representation, and the resulting values were concatenated. Thus the input data was organized in the form of 24+7+10=41 binary values. Correspondingly, the output is encoded as an array of 10 values which, being the output of a softmax layer, encode the probabilities of the next activity. 

\medskip

\noindent{\bf Modeling the deployment times.} Our objective is to model the data available for a collaborative / local model at various moments in time. As we discussed in Section~\ref{sec:Data}, for the collaborative learning models, this depends on the deployment schedule. The CASAS datasets were collected over many years, at time points separated by large intervals, sometimes from successive inhabitants from the same home. To model our scenario, we changed the starting times of the individual datasets to represent different smart environments deployed over the course of 2 months, with a Poisson arrival distribution. This is a realistic model of a small scale deployment by a local health care provider. 

\subsection{Training the Activity Predictor}

Using the preprocessed datasets described in the previous section, we trained the LSTM-based activity predictor from Fig.~\ref{fig:lstm}. For the local training case, we trained 30 different predictors on their local data, for every day of operation. For the centralized predictor, we trained a single shared predictor on the data available, described by Eq.~\ref{eq:training-data-centralized}. For the local and centralized learning, we used the Keras library on top of Tensorflow 2.1.0. For the federated learning, we have trained local predictors with the appropriate local data and updated the shared model using Tensorflow-Federated 0.13.1.

The training configurations for the different models are shown in Table~\ref{tab:hyperparameters}. The code is also available on { https://github.com/sharare90/Privacy-Preserving-Learning}

\begin{table}
    \caption{Local, centralized, and federated training hyperparameters.}
    \centering
    \begin{tabular}{|c|c|c|c|}
    \hline
        \bf Hyperparameters &  \bf Local & \bf Centralized & \bf Federated\\
        \hline
        batch size & 64 & 64  & 64 \\
        % \hline
        number of epochs & 500 & 500 & - \\
                % \hline
        lstm use bias & true & true & true \\
        % \hline
        early stopping patience & 50 & 50 & -\\
        % \hline
        early stopping minimum delta & 0.01 & 0.01 & -\\
        % \hline
        number of rounds & - & - & 20 \\
        % \hline
        % compile learning rate & 0.001 & 0.001 & - \\
        % \hline
        client learning rate & 0.001 & - & 0.001 \\
        % \hline
        server learning rate & - & 0.001 & 0.5 \\
        % \hline
        % compile optimizer & Adam & Adam & - \\
        % \hline
        client optimizer & Adam & - & Adam \\
        % \hline
        server optimizer & - & Adam & SGD \\

        \hline
        
    \end{tabular}
    \vspace{-2mm}
    \label{tab:hyperparameters}
\end{table}

\subsection{Results: Accuracy, Crossover Point and Regret}

The approach we took in this paper is to focus on the  individual user of the specific smart environment. The centralized and federated approaches are not a goal in themselves, they are useful only inasmuch as they help the individual. 

Thus, our performance evaluation is based on measuring the accuracy of the learned predictor (one per home for the local, a shared one for centralized and federated models) on the {\em individual user's data}. As a note, even when the predictor is shared, it will give different results for the individual users. 

We found that the temporal evolution of the accuracy curves fall into several different patterns. Fig.~\ref{fig:evaluation} shows a selection of 12 out of the 30 homes in our dataset, chosen to be representative of the different patterns. Note that the starting day on the $x$ axis varies reflecting the deployment day of the various smart environments. 

We can make several observations:

\noindent{\em Relatively good prediction results.} In interpreting Fig.~\ref{fig:evaluation} we need to keep in mind that the accuracy of random prediction would be 0.1. Fully accurate prediction is not possible, as the users behavior can vary randomly from day to day. The ability to predict the next action with about 45\% accuracy out of 10 possibilities is helpful for many applications for the smart environments. 

\medskip

\noindent{\em The gap between the federated and centralized training is minor.} As expected, we found that the centralized training gives better accuracy results than the federated. However, the differences are small and usually diminish in time. The practical conclusion, for a deployment is that a privacy-aware smart environment would participate in a federated training based collaborative model, as the privacy benefits are significant and the accuracy cost minor.

\medskip

\noindent{\em The crossover point is sometimes very early.} The next question we need to investigate is the relationship between the local and the federated training models. Fig.~\ref{fig:evaluation} shows with a red triangle the crossover point when the local training overtakes the federated learning (if such a point exist in the time interval considered), and with a filled with yellow fill the area corresponding the regret - the accuracy lost if the environment would choose not to participate at all in the federated learning pool. 

We found that the result validate our expectations about the shapes of the accuracy curves: the local learning starts out lower but eventually overtakes the federated learning in 10 out of 12 cases in the figure. In Home 3 the local training starts out better and stays as such, so the regret is zero. In Home 14, where the trends are as expected but the local learning did not yet overtake the accuracy of the federated at the end of the data collection. 

Overall, the location of the crossover point varies. For homes 3, 25, 27 and 29, the crossover happens so early, and the regret area is so small that the deployment of the collaborative learning does not appear to be justified. For other homes, such as Home 2, the crossover happens almost a month after the deployment and the regret is significant. 

%%%%% ---- Lotzi here ---

\subsection{Predicting the Benefits of Federated Training}

\begin{figure}
    \centering
    \includegraphics[width=0.3\linewidth]{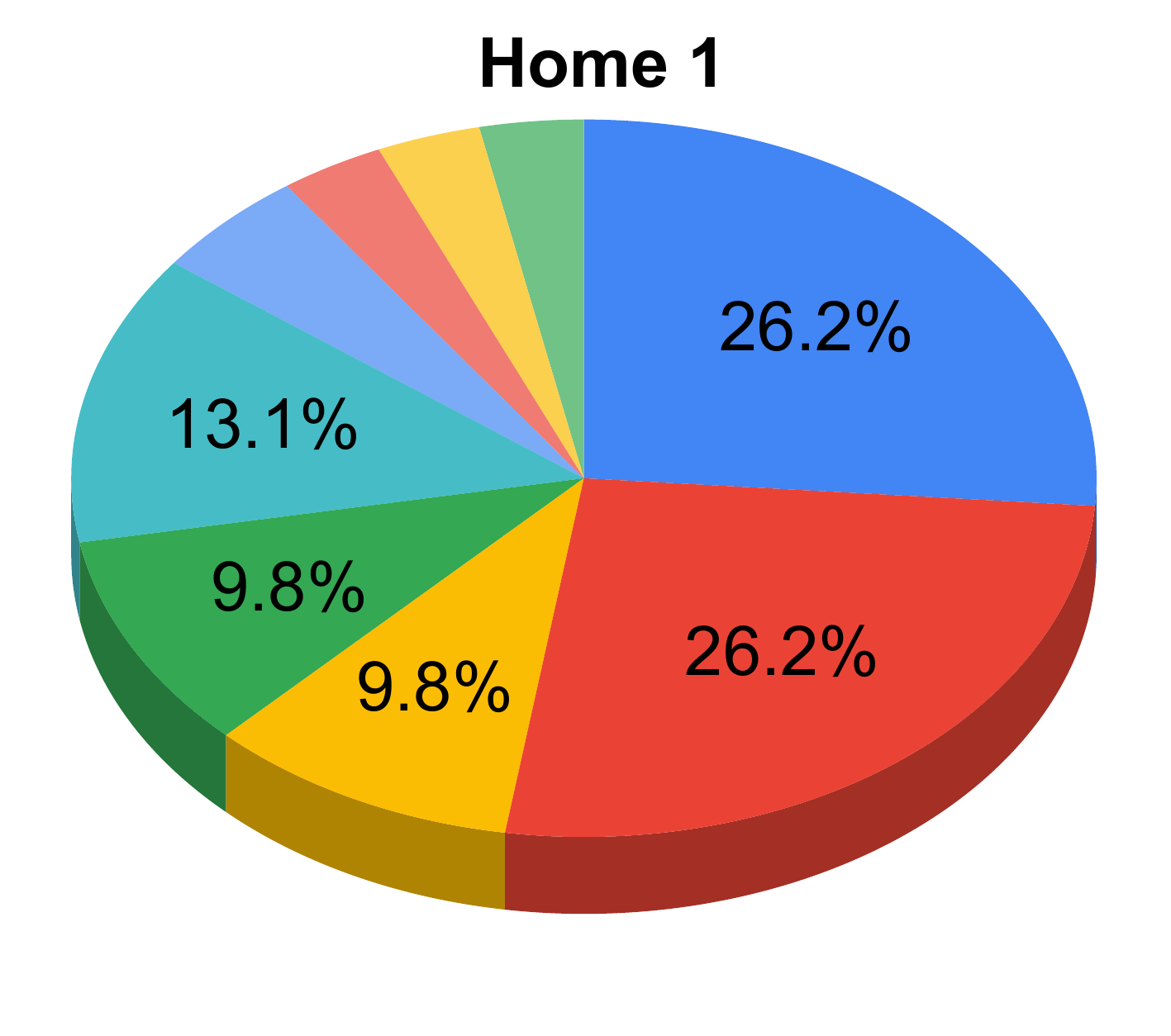}
    \includegraphics[width=0.3\linewidth]{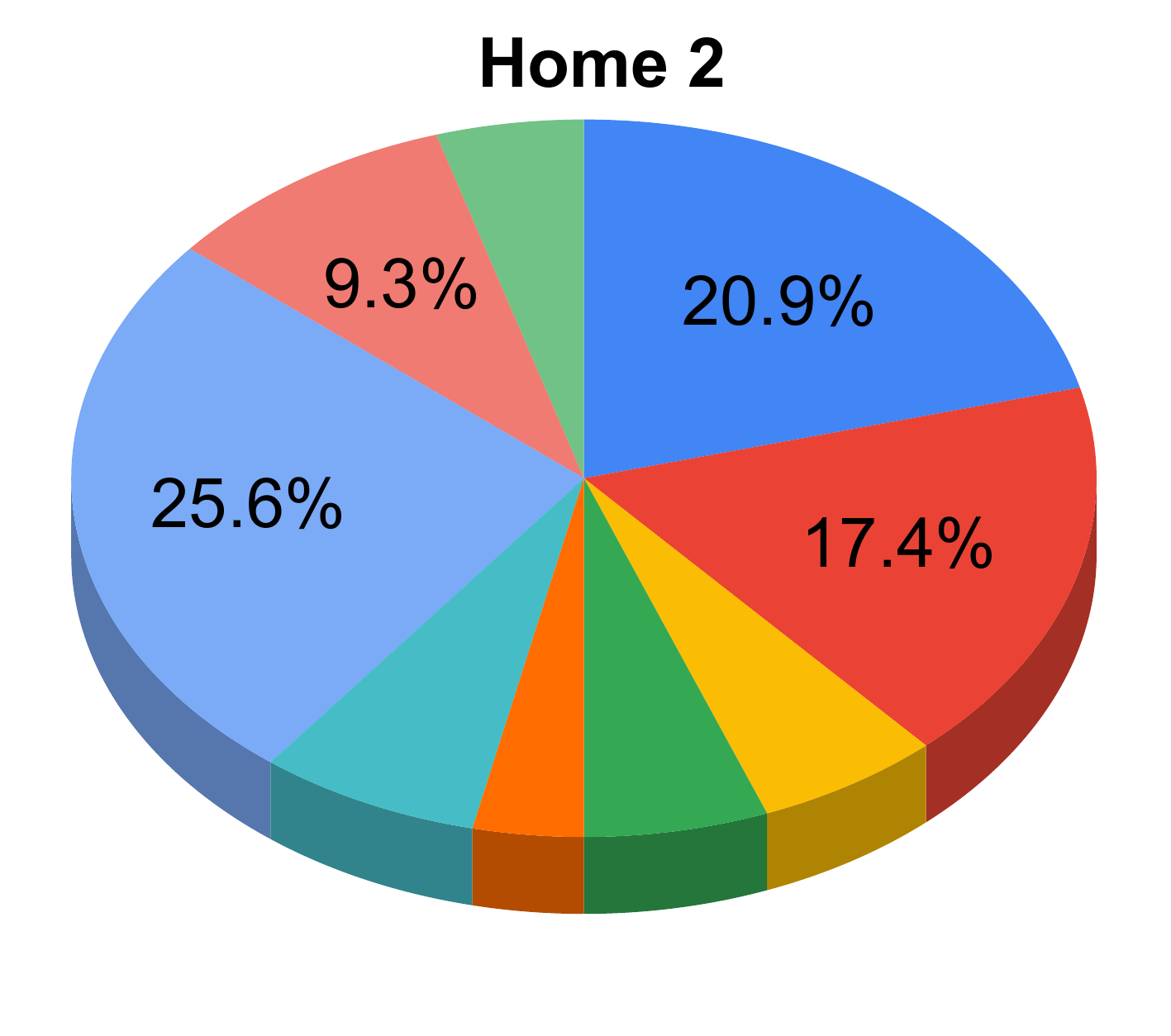}
    \includegraphics[width=0.3\linewidth]{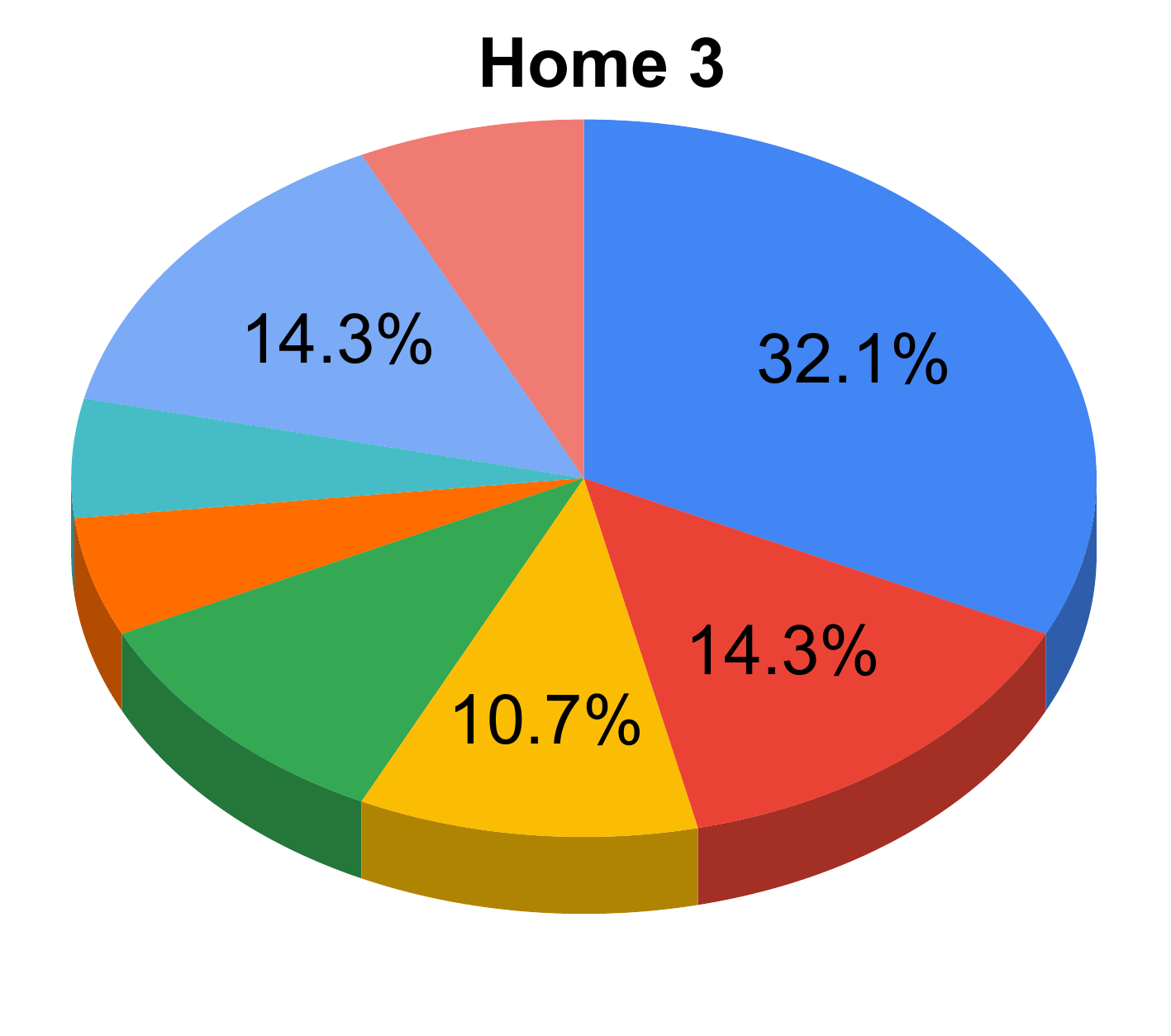} \\
    \includegraphics[width=0.3\linewidth]{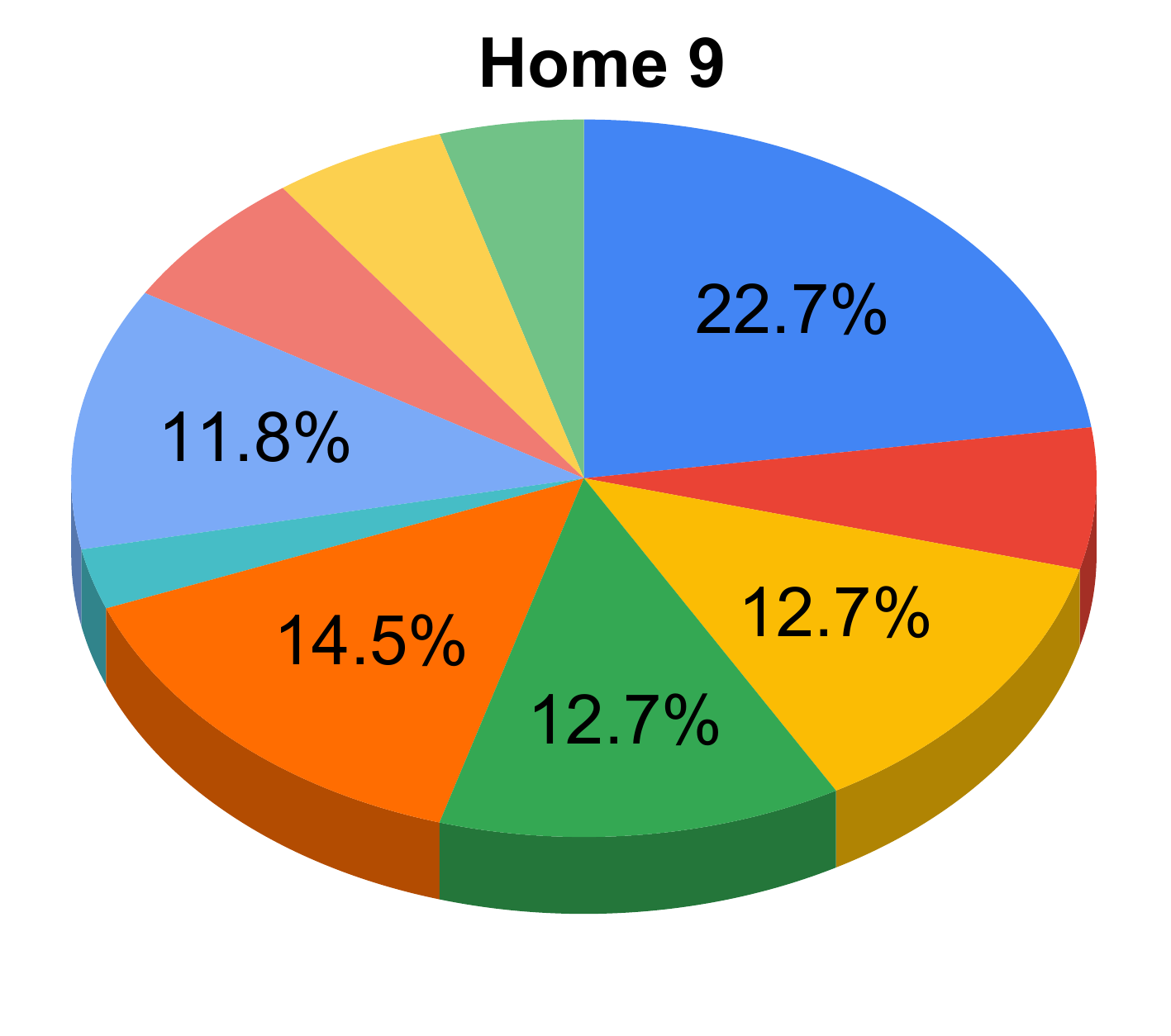} 
    \includegraphics[width=0.3\linewidth]{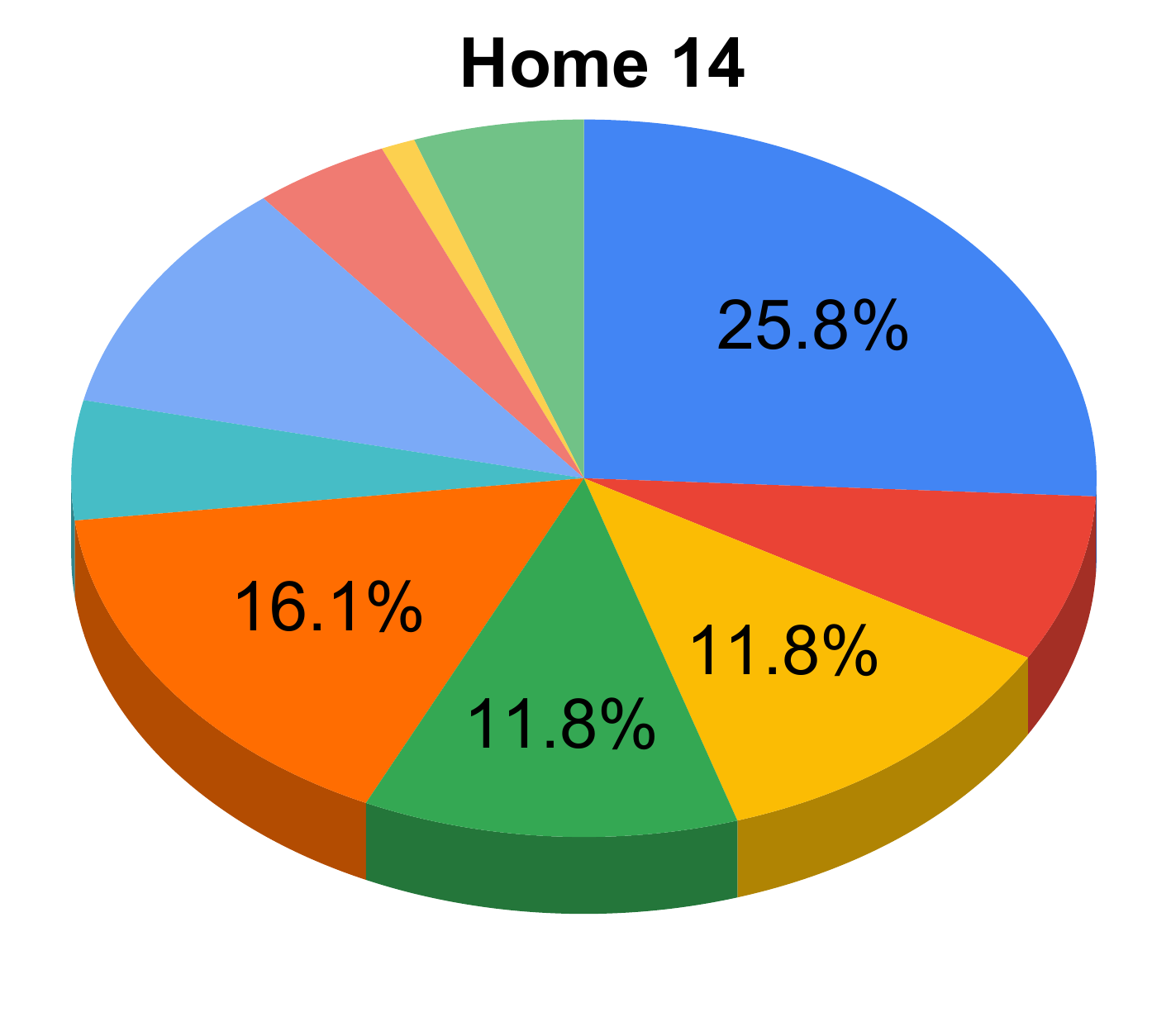}
    \includegraphics[width=0.3\linewidth]{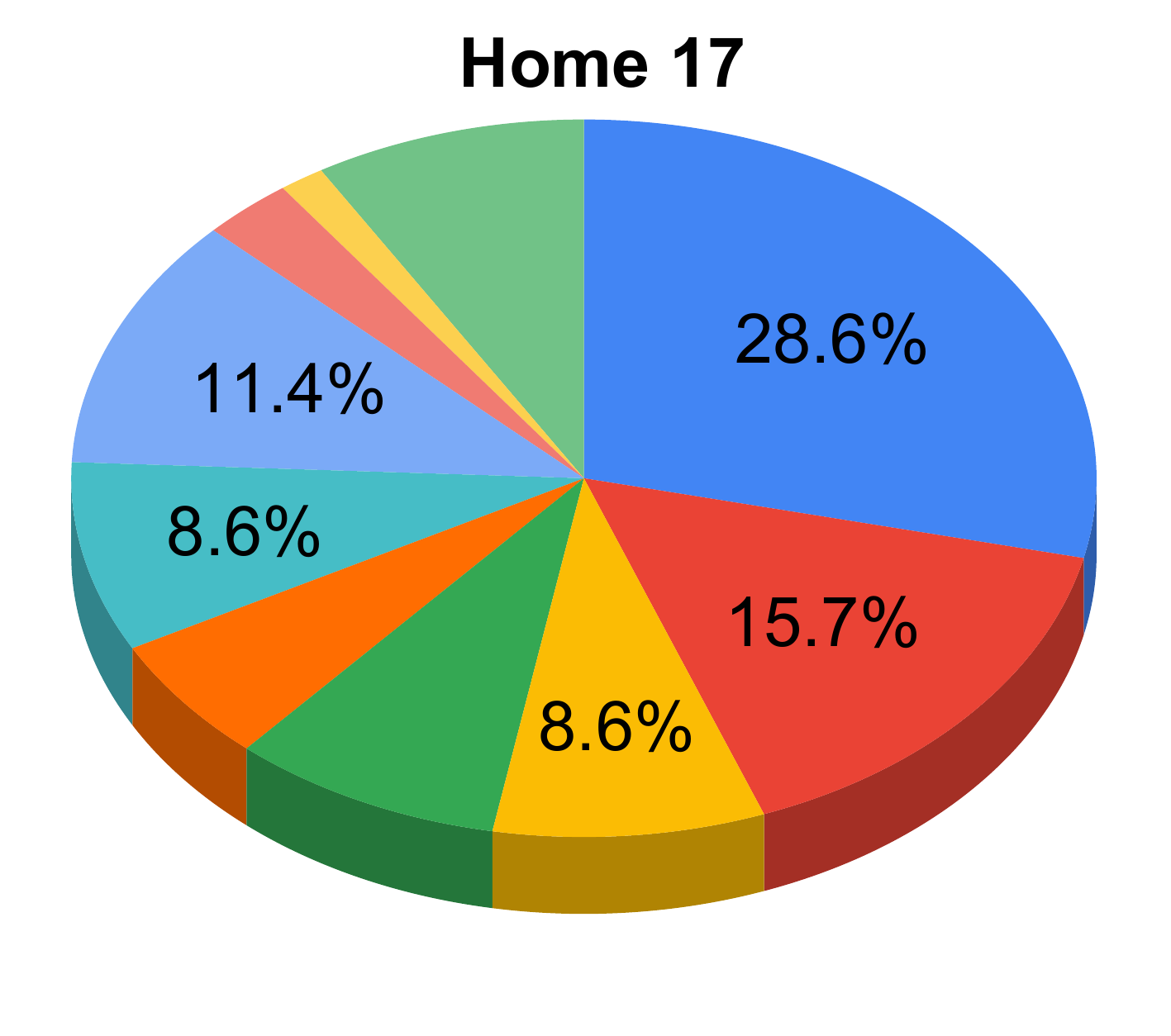} \\
    \includegraphics[width=0.3\linewidth]{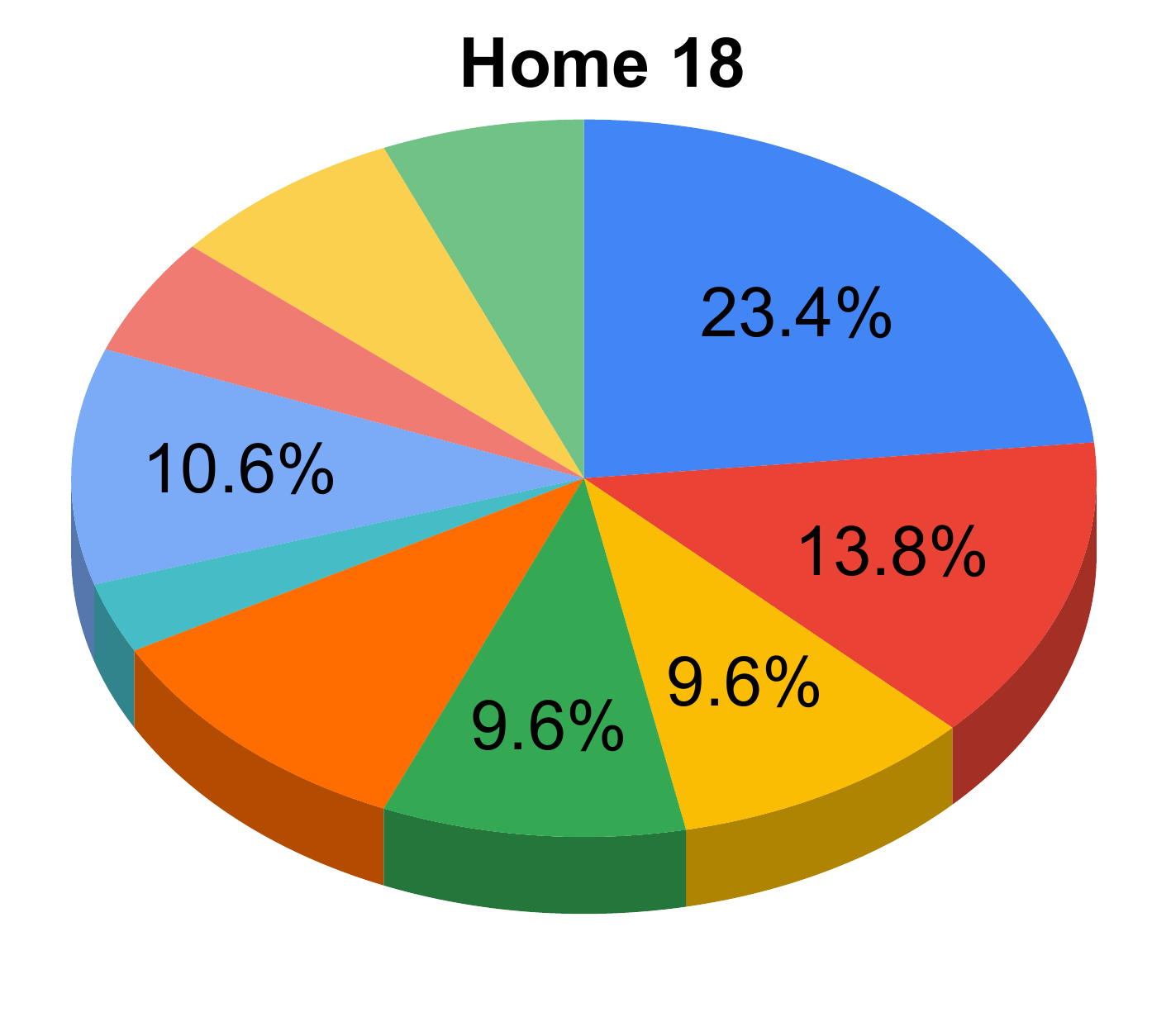}
    \includegraphics[width=0.3\linewidth]{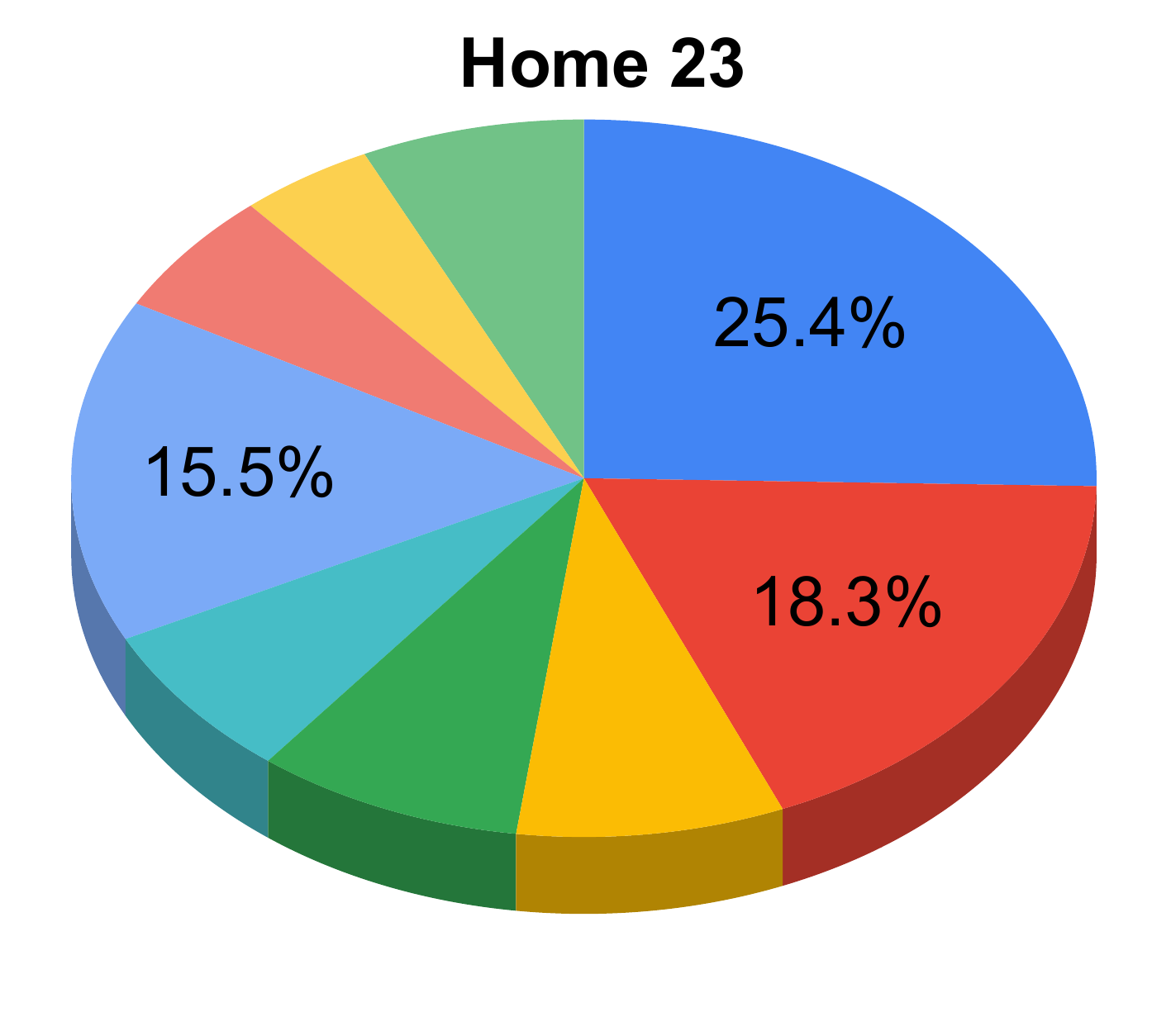} 
    \includegraphics[width=0.3\linewidth]{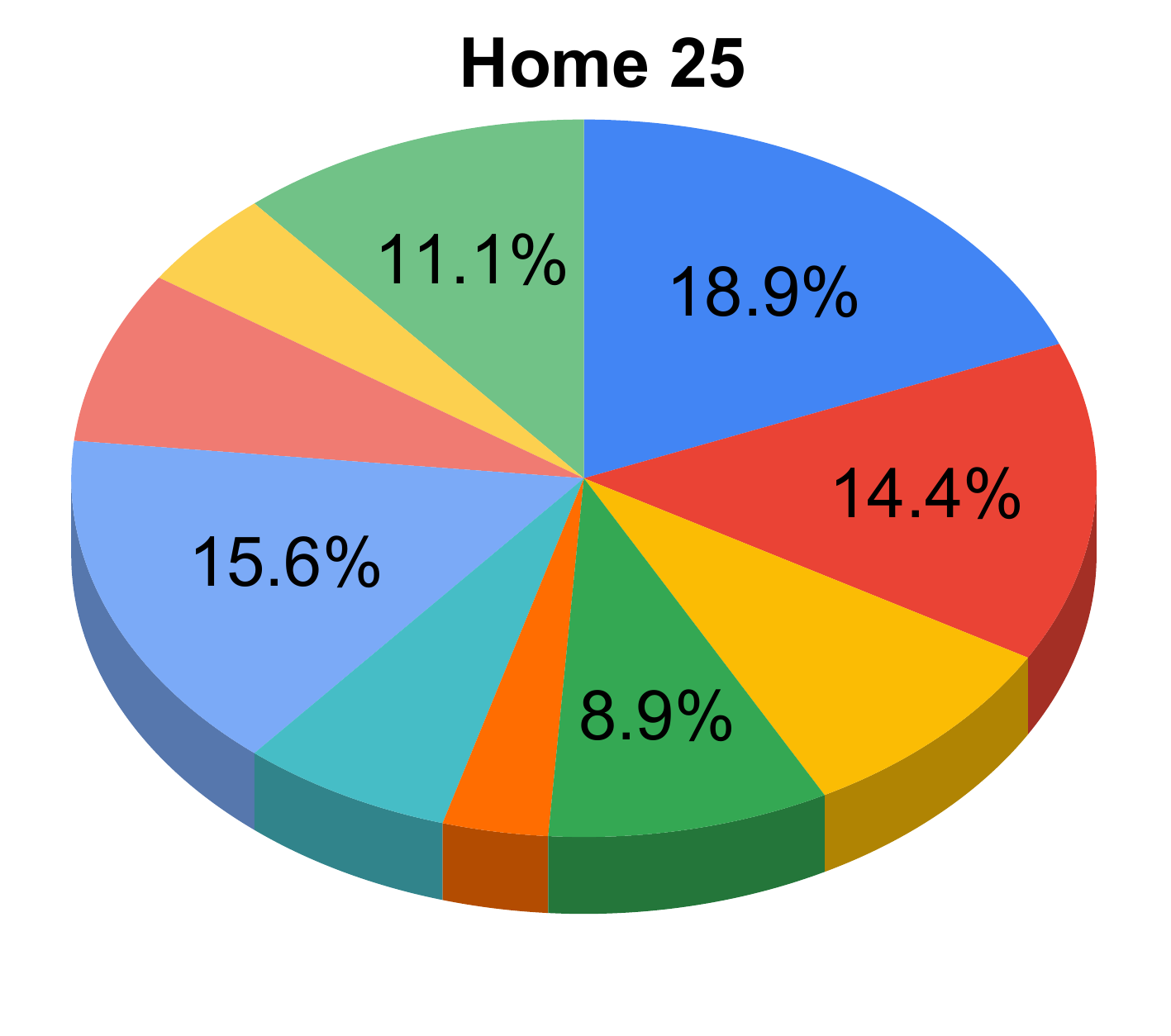} \\
    \includegraphics[width=0.3\linewidth]{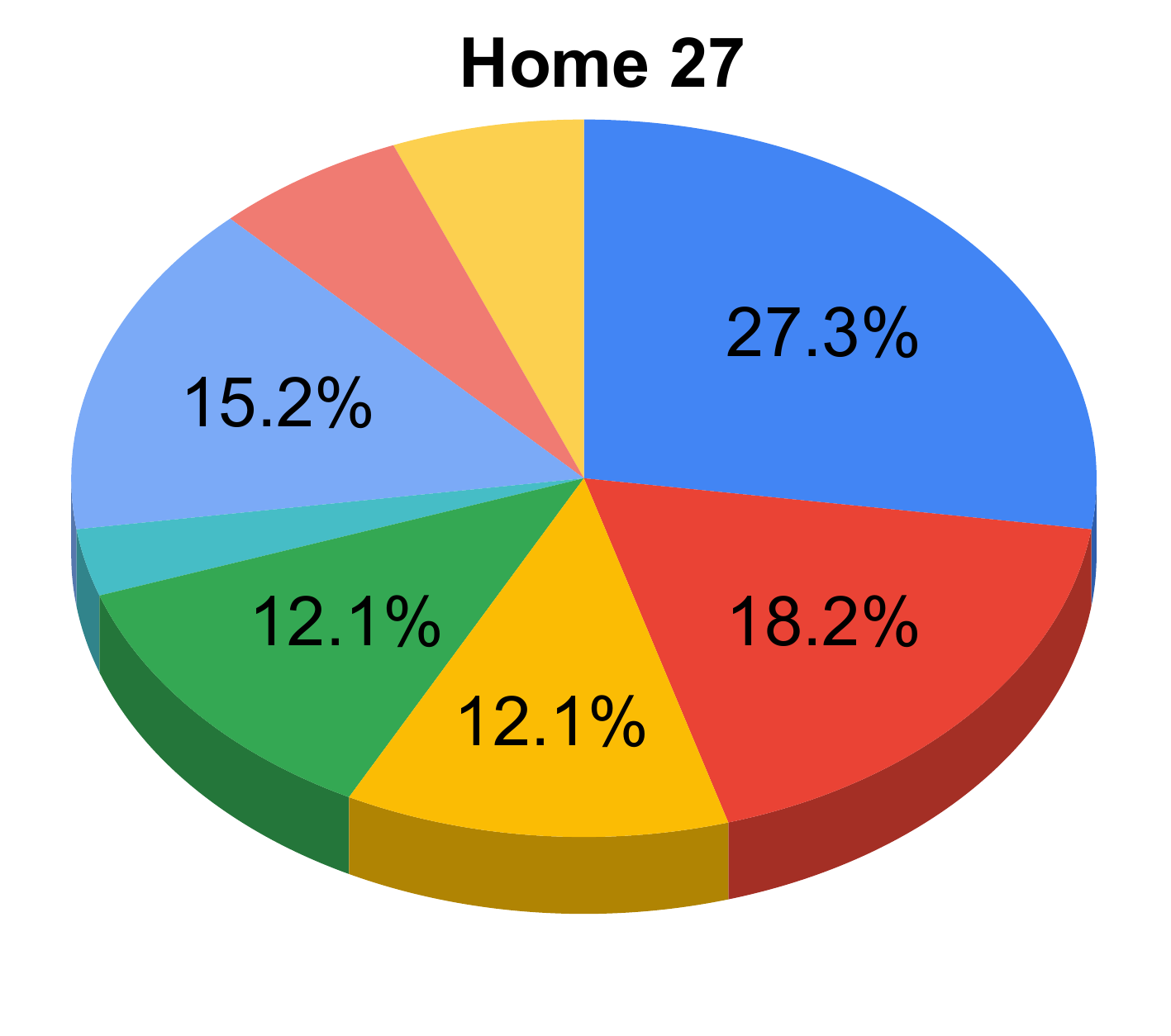} 
    \includegraphics[width=0.3\linewidth]{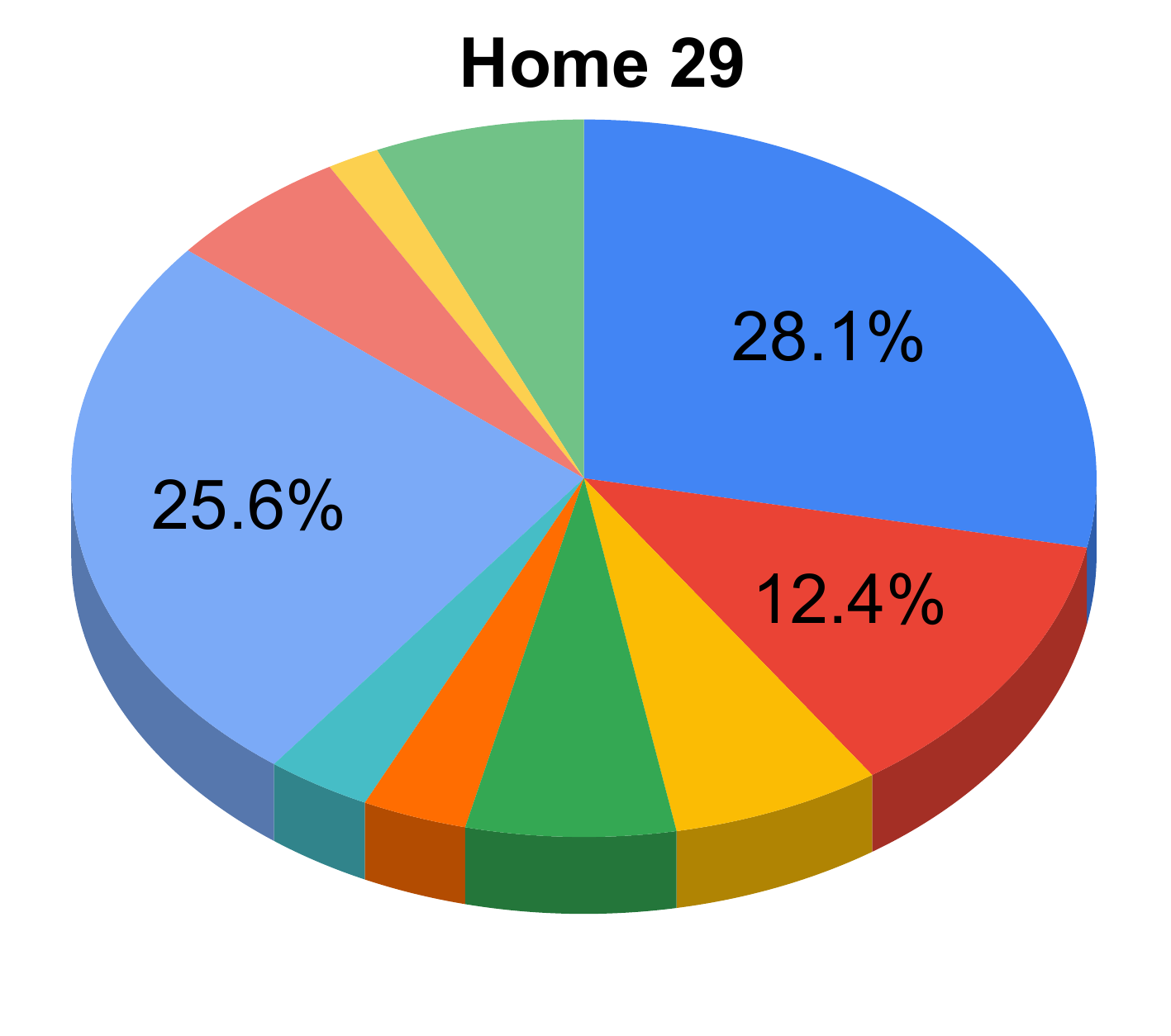}
    \includegraphics[width=0.3\linewidth]{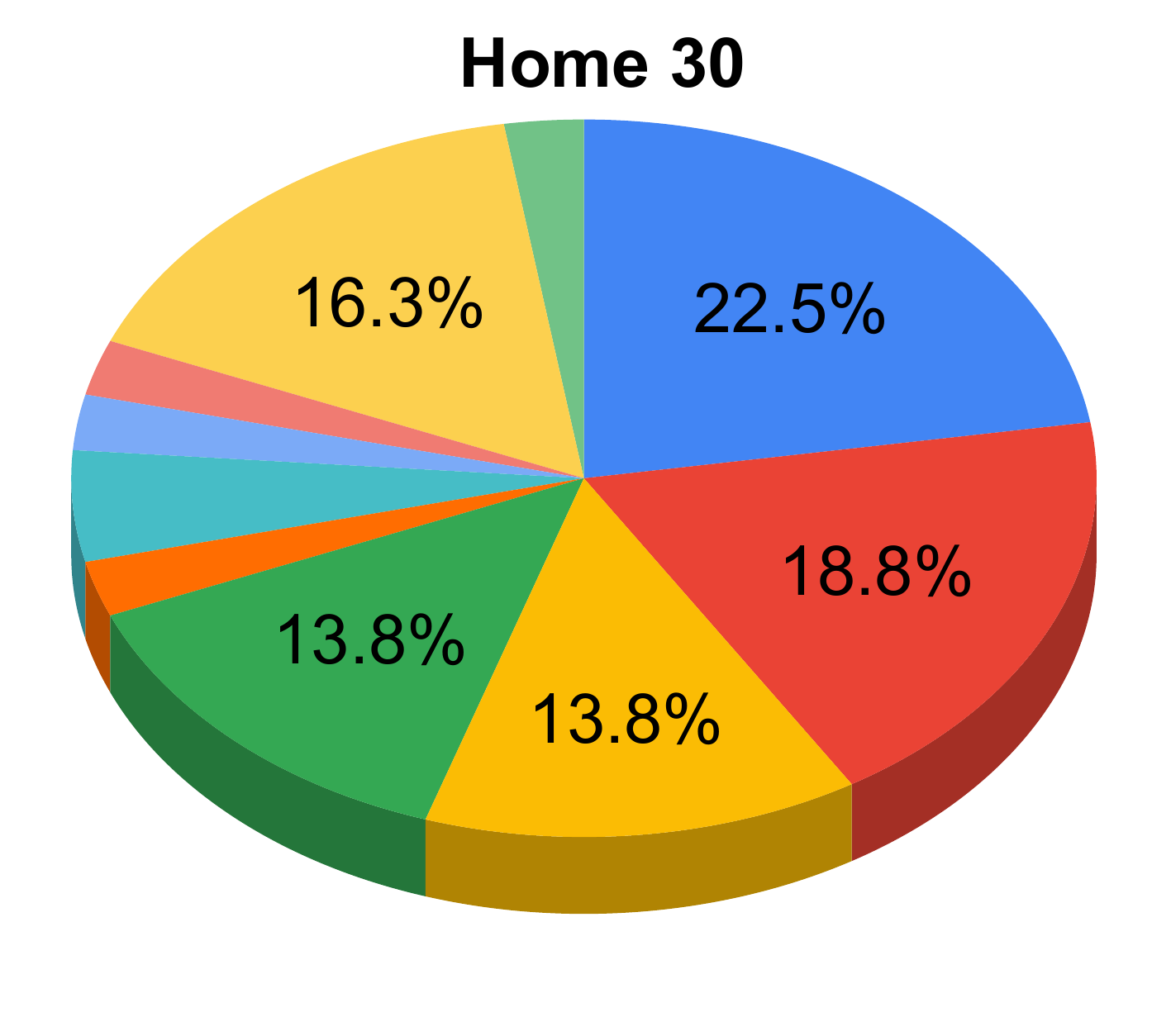} \\
    \includegraphics[width=0.7\linewidth]{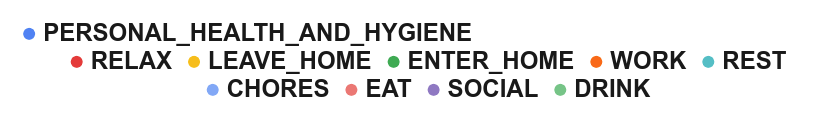}

    \caption{Activity proportions for the same set of homes as shown in Fig~\ref{fig:evaluation}.}
    \label{fig:stats}
    \vspace{-2mm}
\end{figure}

The different profiles found in Fig~\ref{fig:evaluation} raise an interesting question: what factors make for certain homes local learning techniques with very low amount of training data overtake federated learning despite the much more data available to the latter? The main variable here is the way in which the shared model applies to the activities of the specific user. That is: it is not that the local learning is particularly efficient for these users, but that the shared model is not very good for them. The limiting factor of learning the shared model is that the user's behavior is not independent and identically distributed (iid). 

Can we predict the future performance of the federated learning for a given node without joining the pool at all? Our hypothesis is that the proportion of the activities in the home for the first two days contain sufficient information to predict the performance of the predictor. Fig~\ref{fig:stats} shows the activity proportions for the same set of users as shown in Fig~\ref{fig:evaluation}. A visual analysis shows that the users clearly spend different fractions of their time at different activities. 

To predict the relative performance of federated and local training, we trained a classifier whose inputs are the values from Fig~\ref{fig:stats} encoded as floating point values in the $[0,1]$ range, 
% added part
the deployment day, and a day $d \in \{1, 2, \ldots, 45\}$, 
and the output is a single binary value that answers the question:
% ``is the crossover point happens earlier than 7 days from the deployment''?
``is the crossover point happens on day $d$ from the deployment''?
We chose the values of $d$ to be between $1$ and $45$ since if crossover point happens more than 45 days from deployment, we assume that the home will benefit from collaborative learning.
As the number of training data points is small, 
% (one data point per home)
this problem is better suitable for the traditional machine learning approaches. To find the best model, we trained four different classifiers based on decision trees, support vector machines, nearest neighbors and random forests. 

The F1-score of the results are shown in Table~\ref{tab:classifier}. All models achieve a good predictive value, considering the very small amount of training data. The best performing model was the k-nearest neighbor with $k=2$, possibly due to the fact that the best predictor of high federated learning performance is the similarity in profile to nodes already in the pool. 

Overall, the performance of the classifier is sufficient to serve as a decision making aid in helping the user join the federated learning pool or not. One drawback of the approach is that the training of the classifier requires information from all nodes, and thus it can only be done by a centralized authority.

\begin{table}
    %\caption{F1-score results of predicting if smart homes benefit from participation in collaborative learning by using decision tree, SVM, nearest neighbors, and random forest classifiers. }
    \caption{Comparison of classifiers for predicting the benefit of collaborative learning}
    \centering
    \begin{tabular}{|>{\centering\arraybackslash}p{4.0cm}|>{\centering\arraybackslash}p{2.1cm}|}
    \hline
      \bf Classifiers  & \bf F1-score mean \\
      \hline
      Decision Tree &   0.70\\
    %   \hline
      SVM &   0.72\\
    %   \hline
      Nearest Neighbors (k=2)& \bf 0.77\\
    %   \hline
      Random Forest (\# estimators=10)& 0.72 \\
      \hline
    \end{tabular}
    \vspace{-4mm}
    \label{tab:classifier}
\end{table}

\section{Conclusions}
\label{conclusions}

In this paper we considered techniques for learning a human activity predictor for a smart environment in a realistic scenario where the privacy of the users must be weighted against the advantage offered by cloud based, collaborative learning models. We designed an activity predictor using state-of-the-art deep recurrent neural networks and trained it in three separate training scenarios: local, centralized and federated. A novel aspect of our work is that in contrast to previous studies we carefully accounted for what training data is available for the environments at every point in time. Our experiments had shown that there is only a minor difference between the centralized and federated approach, thus the greater privacy of federated learning would make it the preferred cloud based model. Furthermore, our experiments had also shown that the local training model will overtake the accuracy of the federated model for almost all the cases. In fact, for a significant subset of the environments, this crossover points happens within a couple of days of the deployment. To allow the user to predict this and use it to maximize his privacy, we trained a classifier that can predict the early crossover based on the first days' data, with no disclosed information. 

\smallskip 

\noindent {\bf Acknowledgment:} This work was supported by the National Science Foundation under Grant No. 1800961. 

%\begin{IEEEkeywords}
%\end{IEEEkeywords}

\clearpage

\bibliographystyle{IEEEtran}
% \bibliography{./bibliography/IEEEabrv,./bibliography/IEEEexample}
\bibliography{references.bib}

\end{document}